\documentclass{article}
\usepackage{iclr2027_arxiv,times}

\usepackage{amsmath,amsfonts,bm}

\def\eqref#1{equation~\ref{#1}}

\def\1{\bm{1}}

\DeclareMathAlphabet{\mathsfit}{\encodingdefault}{\sfdefault}{m}{sl}
\SetMathAlphabet{\mathsfit}{bold}{\encodingdefault}{\sfdefault}{bx}{n}

\usepackage{hyperref}
\hypersetup{hidelinks}
\usepackage{url}
\usepackage{graphicx}
\usepackage{booktabs}
\usepackage{float}
\usepackage{multirow}
\usepackage{xcolor}
\usepackage{amsmath}

\newcommand{\tb}[1]{{\bfseries\boldmath #1}}

\newcommand{\opd}{OPD}
\newcommand{\sft}{SFT}
\newcommand{\chartqa}{ChartQA}
\newcommand{\geo}{Geometry3K}

\title{Do As I Do, Not As I Say: \\
On-Policy Distillation Inherits the Teacher in Vision-Language Models}

\iclrfinalcopy

\author{Frank F. Yang\thanks{All four authors contributed equally to this work.}
\quad Kefei Yan\footnotemark[1]
\quad Yiren Wang\footnotemark[1]
\quad Yiyang Wang\footnotemark[1] \\
Georgia Institute of Technology \\
Atlanta, GA, USA \\
\texttt{\{frank.yang,kyan66,ywang4361,ywang3420\}@gatech.edu}
}

\begin{document}

\maketitle
\begin{abstract}
Improving the reasoning capabilities of vision-language models~(VLMs) often relies on post-training with supervision from stronger models. 
Two common strategies are supervised fine-tuning (SFT), which trains on teacher-generated demonstrations, and on-policy distillation (OPD), which provides teacher feedback on trajectories generated by the student itself.
While OPD can be more data-efficient by supervising student-generated trajectories, it is unclear \textit{how the teacher's supervision at each state affects the student's learned behavior}.
To study this question, we conduct a controlled comparison of OPD and filtered SFT across data budgets and tasks where the teacher exhibits different levels of reliability, and analyze how these differences shape student accuracy and completion behavior. 
Across the two tasks, we find that OPD is effective only when the teacher remains reliable on student-generated trajectories. 
On ChartQA, where the teacher completes 99\% of its solutions, OPD trained on only 10\% of the data comes within 1.4 accuracy points of full-data SFT. 
In contrast, on Geometry3K, where the teacher fails to complete 22\% of its solutions, OPD underperforms SFT by 7.9 points, with most of the gap explained by incomplete responses. 
More strikingly, starting OPD from an SFT student reduces its completion rate from 76\% to 62\%, effectively erasing the completion behavior learned through SFT. 
Filtering teacher demonstrations for completeness alone recovers much of SFT's advantage. 
These results show that on-policy supervision can transfer not only teacher knowledge but also persistent failure modes on student-visited states. Code is available at \url{https://github.com/ffy208/opd-inherits-teacher-vlm}.
\end{abstract}

\section{Introduction}

Distilling a large vision-language model (VLM) into a small one is the standard way to make multimodal
reasoning cheap enough to deploy. The common recipe is supervised fine-tuning on teacher-written solutions
\citep{magister2023teaching,ho2023reasoningteachers} (\sft), which is behavior cloning: the student sees only states the teacher visited, drifts into others at
inference, and its errors compound with sequence length \citep{ross2010efficient,ross2011dagger}. On-policy distillation (\opd)
is the DAgger-style remedy \citep{agarwal2024gkd,gu2024minillm}: the student samples its own solution, the
teacher scores every generated token, and the student minimises a per-token divergence to the teacher on the
states it actually reaches. For text-only language models, on-policy distillation on a twentieth of the training set has been
reported to outperform supervised distillation on the full set \citep{agarwal2024gkd}, and it is now a standard
post-training stage \citep{yang2025qwen3,lu2025onpolicy}. A dense 2026 literature now adapts the objective to VLMs by re-weighting the teacher's
signal toward visually grounded tokens \citep{liu2026vaopd,yoon2026decomposed,xue2026fpopd}.

While these methods change the objective and report the gains of the change, they leave open the question that
comes first in practice: given a teacher and a few hundred to a few thousand training questions, when does plain
on-policy feedback improve on filtered teacher demonstrations, and which behavior of the teacher does the
student acquire along the way? Related work observes that teacher feedback degrades on flawed student prefixes
\citep{qiu2026opdaha,sun2026sopd}; our experiments add completion as the behavior at stake, a matched
comparison with filtered demonstrations, and a student that loses the completion gain it already had.

The two recipes differ in cost and in what they transfer: \opd{} keeps the teacher in the loop and
supervises the student on its own states with whatever the teacher does there; \sft{} pays once for teacher
solutions and transfers only those that survive a correctness filter. Which difference dominates depends on the
task, and in our experiments the wrong choice costs up to $7.9$ points. Accuracy at a single budget cannot tell
whether a gap comes from the amount of data or the number of training steps, from reasoning errors or from solutions
that were never completed, from the objective or from the filter.

To investigate these questions, we conducted (Section~\ref{sec:method}) a controlled comparison of the two channels across nested
data budgets with each channel's update count fixed, token-role divergence logging at every training step,
and completion diagnostics, applied to a Qwen3-VL 2B student and 8B teacher \citep{bai2025qwen3vl} on
\chartqa{} \citep{masry2022chartqa} and \geo{} \citep{lu2021intergps}.

Our contributions are threefold:
\begin{itemize}
    \item \textbf{A controlled empirical comparison of OPD and
    filtered SFT.}
    We systematically evaluate both training recipes across
    four data budgets, four seeds, and two vision-language
    reasoning tasks with different teacher completion rates.
    We further examine the sensitivity of their relative
    performance to SFT training length and evaluation
    distribution.

    \item \textbf{Identification of a completion-specific
    failure mode.}
    We show that final-answer accuracy can obscure substantial
    differences in completion behavior. By decomposing
    accuracy differences according to whether each method
    produces a final answer, we characterize how incomplete
    responses contribute to the performance gap. We further
    demonstrate that continued OPD can undo completion
    behavior previously acquired through SFT.

    \item \textbf{An empirical investigation of the failure.}
    We examine teacher behavior on student-generated prefixes,
    compare demonstrations filtered by correctness and
    completeness, and evaluate five OPD training variants.
    These analyses characterize the persistence of the
    completion failure and investigate the possible role of
    teacher feedback on student-visited states.
\end{itemize}
Two findings stand out.

\begin{itemize}
  \item \textbf{A persistent failure to complete solutions.} On \geo, where the teacher itself does not
  finish 22\% of its solutions within the token budget, \opd{} students at 900 problems conclude 63\% of test
  responses against 76\% for \sft{} and trail \sft{} by $7.9$ points ($-10.5$ to $-5.3$), of which $6.6$ points come from problems that only one method concludes. \opd{} started from the \sft{} student returns its rollout termination rate to plain \opd's within 40 steps and ends at
  62\% test completion, the zero-shot level.
  \item \textbf{A small, non-robust data-efficiency advantage with a reliable teacher.} On \chartqa, \opd{}
  with 300 training questions reaches $0.826$ accuracy, within $1.4$ points of \sft{} trained on 3{,}000
  ($0.840$), and leads \sft{} by $2.7$ points at the same budget under the protocol's fixed \sft{} length (per
  seed $+0.6$ to $+5.8$) and by $1.1$ to $1.5$ points, with intervals that include zero, against an \sft{}
  whose training length is tuned.
\end{itemize}

Together, these results characterize a previously underexplored failure mode of on-policy distillation and give a working rule and a target: on these tasks \opd{} was competitive with filtered \sft{} at small
budgets only where the teacher reliably completed its solutions; where it did not, the student's completion
failures concentrated on the problems the teacher does not finish, and from those prefixes the teacher
rarely concludes, even with the answer in its prompt. Whether that causes the failure or accompanies it awaits
a teacher trained to conclude from such states (Section~\ref{sec:geo}).
Our findings motivate evaluating completion alongside accuracy and examining the reliability of teacher feedback on student-visited states.
\section{Method: a controlled comparison of two supervision channels}
\label{sec:method}

\subsection{Two supervision channels}
\label{sec:channels}

Let $x$ be a prompt (an image and a question), $\pi_T$ the teacher and $\pi_\theta$ the student
\citep[distillation as matching the teacher's outputs,][]{hinton2015distilling}. \sft{} samples solutions
$y \sim \pi_T(\cdot \mid x)$ once, keeps those whose final answer is correct \citep{ho2023reasoningteachers,yuan2023rft},
and minimises the token-level cross-entropy $-\sum_t \log \pi_\theta(y_t \mid x, y_{<t})$ on the kept set. \opd{}
instead samples a batch of rollouts $y^{(i)} \sim \pi_\theta(\cdot \mid x^{(i)})$ at every step and minimises the
mean, over all generated tokens of the batch, of a divergence between the two next-token distributions on the
student's own prefix,
\begin{equation}
  \mathcal{L}(\theta) = \frac{1}{\sum_i |y^{(i)}|}\sum_{i}\sum_{t=1}^{|y^{(i)}|} D\big(\pi_\theta(\cdot \mid x^{(i)}, y^{(i)}_{<t}) \,\|\, \pi_T(\cdot \mid x^{(i)}, y^{(i)}_{<t})\big),
  \label{eq:opd}
\end{equation}
with $D$ the reverse KL by default \citep{agarwal2024gkd,gu2024minillm}. The divergence is taken over the full
vocabulary, including the end-of-turn token, at every generated position up to and including the first
end-of-turn token or the generation budget; a rollout that runs to the budget contributes every one of its
positions.

The case for on-policy supervision is DAgger's \citep{ross2010efficient,ross2011dagger}: querying the expert on
the learner's own states removes the compounding error of behavior cloning, provided the expert's action on
those states is worth imitating. The two channels therefore differ in two things at once: the states on which
the teacher is queried (its own prefixes against the student's) and the teacher behavior that reaches the
student (only correct solutions against everything the teacher does). Our design separates these.

\subsection{Design: what is held fixed and what varies}
\label{sec:design}

Rather than propose a new objective, we hold everything but the supervision channel and the amount of data
fixed: one frozen teacher and one student initialisation, with LoRA on the language-model projections and the
vision encoder frozen, so the channels differ in language-side supervision and not in visual representation;
one prompt, answer format, generation budget, scorer, and held-out test set per task; budgets that are nested
prefixes of one seeded shuffle, so both channels see the same questions and a larger budget adds questions
rather than replacing them; and a number of optimizer updates that is fixed within each channel whatever the
budget, so that along a curve only the amount of data changes (the counts are not matched between channels; a
training-length control on \sft{} checks this). The teacher is neither prompted for completeness nor retrained:
its reliability varies only through the task, and two tasks on which the same teacher is reliable and
unreliable turn that dependence into a difference between tasks rather than between teachers.

The channel (filtered demonstrations against on-policy token-level feedback) is crossed with four budgets, two
tasks, and four paired seeds. Because the channel bundles the state distribution with the filter, two further
\sft{} data sets (no filter; every teacher sample with an answer line, correct or not) separate the two. Five
single-component interventions on \opd{} (the loss on terminated rollouts only, the teacher's sampling
temperature, the forward KL, a sequence-mean loss, initialisation from the \sft{} student) change one part of
the recipe each, and a larger teacher and two out-of-distribution test sets check the dependence on teacher size
and test distribution.

\subsection{What we measure}
\label{sec:measures}

Beyond accuracy, three complementary analyses characterize the observed differences and probe possible explanations. \emph{Completion diagnostics}: a response
is \emph{complete} if it contains an answer line (one without counts as wrong), \emph{terminates} if it emits
the end-of-turn token before the budget, and is \emph{truncated} if it reaches the budget; paired accuracy
differences are decomposed by which method concluded the problem. \emph{Feedback logging}: at every \opd{} step
we record each generated token's divergence and a heuristic role (\emph{answer line}, \emph{numeric value},
\emph{arithmetic}, \emph{text}); a role's \emph{KL concentration} is its share of the batch's divergence divided
by its share of tokens. \emph{Teacher probes}: the teacher is continued greedily from prefixes of the student's
truncated and concluded outputs and its end-of-turn probability is read along them, to ask what it does on the
student's states rather than on its own.

\section{Experimental setup}
\label{sec:setup}

\paragraph{Models and tasks.}
The student is Qwen3-VL-2B-Instruct, trained with LoRA \citep{hu2022lora} of rank 64 on the language-model
projections, and the teacher Qwen3-VL-8B-Instruct \citep{bai2025qwen3vl}, frozen; a 32B teacher is used in
Section~\ref{sec:ood}. \chartqa{} \citep{masry2022chartqa} is
restricted to human-written questions (3{,}000 training and 500 held-out test questions sampled with a fixed
seed), where the 8B teacher answers 84\% correctly and concludes 99\% of its solutions with an answer line.
\geo{} \citep{lu2021intergps} has 2{,}101 training and 500 of the 601 official test problems with numeric
answers, where the teacher answers 57\% correctly and concludes only 78\% of its solutions within the
1{,}536-token budget. The correctness filter of \sft{} keeps 2{,}402 of the 3{,}000 \chartqa{} teacher samples
(80\%) and 1{,}163 of the 2{,}101 \geo{} samples (55\%), so an \sft{} budget of $b$ questions trains on about
$0.8b$ or $0.55b$ demonstrations. Two out-of-distribution chart sets, 461 numeric questions from CharXiv
\citep{wang2024charxiv} and 500 factoid questions from ChartQAPro \citep{masry2025chartqapro}, score every
\chartqa-trained model.

\paragraph{Protocol.}
Every model sees the same prompt: solve step by step and end with a line \texttt{Answer: <value>}. Accuracy is measured using
relaxed match with 5\% numeric tolerance \citep{methani2020plotqa}, the \chartqa{} convention, which we also apply to \geo{} (the \geo{}
ordering is unchanged at 1\% and 0.1\%; Appendix~\ref{app:tolerance}). The stored records predate the evaluator's stop-reason
field, so truncation is inferred by re-tokenising the output (within 8 tokens of the budget;
Appendix~\ref{app:completion}). For every correctness-filtered \sft{} and \opd{} model in the paper fewer than
1\% of responses stop without an answer line, so an incomplete response is a truncated one (exceptions: the two
\sft{} data controls of Section~\ref{sec:geo}, at 14\% and 2\% of responses at 300 problems). \sft{} runs 302 optimizer steps (two epochs of the full data) at every budget;
\opd{} runs 150 steps of 16 rollouts at temperature 1.0 with a 512-token budget on charts and 1{,}536 on
geometry; the step count is fixed within each method, not matched between them. Budgets $b \in \{100, 300, 900,
3000\}$ questions (charts) or $\{100, 300, 900, 2101\}$ problems (geometry) are nested prefixes of a seeded
shuffle. Seeds change the data order, rollout sampling, and LoRA initialisation; the test set is fixed; every
point and every variant has four seeds. Full hyperparameters are in Appendix~\ref{app:hparams}.

\paragraph{Intervals.}
Every interval is a 95\% question-clustered bootstrap interval: the 500 test questions are resampled with
replacement, every seed's outcome on a resampled question is kept together, and differences are paired by
question within each seed and averaged over seeds; this treats the four trained models as fixed. For the
headline comparisons we also give each seed's difference and a $t$ interval over the four seeds, whose target is
the method rather than these models (Appendix~\ref{app:intervals}), and we call a difference robust only if the clustered and the seed-level interval both
exclude zero.

\section{Experiments and Results}

\subsection{With a reliable teacher, \opd{} is competitive at small budgets}
\label{sec:chartqa}

\begin{figure}[t]
  \centering
  \includegraphics[width=\linewidth]{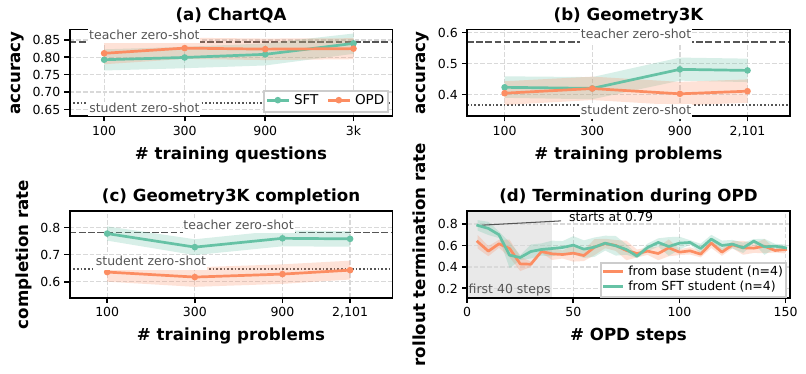}
  \caption{Accuracy against training questions at a fixed number of optimizer updates per method (mean over
  four seeds) on \chartqa{} (a), where the teacher concludes 99\% of its solutions, and \geo{} (b), where it
  concludes 78\%; (c) share of \geo{} test responses with an answer line. Shaded bands: 95\% question-clustered
  bootstrap intervals; grey dashed: zero-shot teacher; grey dotted: zero-shot student. (d) Rollout termination rate
  during \opd{} on \geo{} at 900 problems (five-step windows, mean and $\pm 1$ s.d. over four seeds) from the
  base and from the filtered-\sft{} student, which starts at $0.79$ and is in plain \opd's band within 40 steps
  (shaded).}
  \label{fig:curves}
\end{figure}

\begin{table}[t]
  \centering
  \scriptsize
  \setlength{\tabcolsep}{3pt}
  \caption{\chartqa: accuracy on 500 held-out questions, mean over four seeds (across-seed sd in brackets;
  every mean has a question-clustered 95\% interval of $\pm 0.03$), with the paired \opd{} minus \sft{}
  difference, its clustered interval, and the $t$ interval over the four per-seed differences. The last two
  rows use the \sft{} length that scored best in Appendix~\ref{app:sft-length}. Bold: the higher accuracy
  where the clustered interval excludes zero, and intervals that exclude zero. Zero-shot student $0.668$,
  teacher $0.844$.}
  \label{tab:chartqa}
  \begin{tabular}{lcccc}
    \toprule
    & \multicolumn{2}{c}{Accuracy} & \multicolumn{2}{c}{\opd{} $-$ \sft{}} \\
    \cmidrule(lr){2-3} \cmidrule(lr){4-5}
    Questions & \sft & \opd & paired, clustered 95\% & $t$ over seeds \\
    \midrule
    100 (3\%) & $0.793$ (.022) & $0.811$ (.009) & $+0.018$ [$-0.001$, $+0.038$] & [$-0.006$, $+0.043$] \\
    300 (10\%) & $0.799$ (.015) & \tb{$0.826$} (.007) & \tb{$+0.027$ [$+0.008$, $+0.047$]} & [$-0.008$, $+0.062$] \\
    900 (30\%) & $0.808$ (.011) & $0.823$ (.004) & $+0.015$ [$-0.004$, $+0.035$] & \tb{[$+0.003$, $+0.027$]} \\
    3000 (100\%) & $0.840$ (.007) & $0.825$ (.008) & $-0.015$ [$-0.032$, $+0.002$] & [$-0.033$, $+0.003$] \\
    \midrule
    100, \sft{} 38 steps & $0.797$ (.012) & $0.811$ (.009) & $+0.015$ [$-0.005$, $+0.035$] & \tb{[$+0.001$, $+0.028$]} \\
    300, \sft{} 38 steps & $0.816$ (.005) & $0.826$ (.007) & $+0.011$ [$-0.006$, $+0.028$] & \tb{[$+0.001$, $+0.020$]} \\
    \bottomrule
  \end{tabular}
\end{table}

Table~\ref{tab:chartqa} and Figure~\ref{fig:curves}a give the \chartqa{} curve. \opd's mean is above
\sft's at 100, 300, and 900 questions by $1.5$ to $2.7$ points, and eleven of the twelve per-seed differences are positive, but no lead is robust: the 300-question lead excludes zero
under the clustered interval and not under the seed-level one (per seed $+0.6$ to $+5.8$), the
900-question lead the reverse, and the 100-question lead under neither. With a tenth of the questions, \opd{} at 300 ($0.826$) comes within $1.4$ points of \sft{} on all 3{,}000
($0.840$), a paired difference of $-1.4$ [$-3.1$, $+0.4$]; we did not pre-specify a non-inferiority margin, so
we say \opd{} approaches full-data \sft{} at a data ratio of the kind reported for text-only \opd{}
\citep{agarwal2024gkd} rather than matches it. At
3{,}000 questions the ordering reverses by $1.5$ points, all four seeds negative and the clustered interval
touching zero. The flat \opd{} curve is a property of the fixed-update design: 2{,}400 rollouts are 24 passes
over 100 questions and 0.8 over 3{,}000, so \opd{} gains $0.8$ points per decade of data against $3.1$ for
\sft{} (Appendix~\ref{app:scaling}).

\paragraph{Control: the training length of \sft.}
The protocol's 302 steps are two epochs of the full data but about 20 epochs at 300 questions and 60 at 100, so
the design could be handicapping \sft{} at low budgets. It is: retrained for 19, 38, 75, and 150 steps
(Appendix~\ref{app:sft-length}), \sft{} peaks at 38 steps with $0.816$ at 300 questions against $0.799$, and
against it \opd's lead shrinks to $+1.1$ [$-0.6$, $+2.8$] at 300 questions and $+1.5$ [$-0.5$, $+3.5$] at 100
(Table~\ref{tab:chartqa}): \opd's mean is above every \sft{} length, but its lead over a length-tuned \sft{} is
1 to 1.5 points and not robust, the tuned \sft's seed spread is no larger than \opd's, and the \sft{} length
was selected on the test set, since the protocol has no validation split.

\subsection{The teacher's feedback lands on the answer line first}
\label{sec:dynamics}

\begin{figure}[t]
  \centering
  \includegraphics[width=0.9\linewidth]{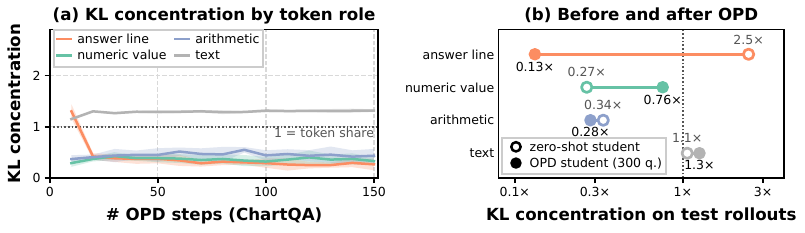}
  \caption{Where the teacher's per-token feedback lands on \chartqa. KL concentration is a token role's share of
  the batch's reverse KL divided by its share of tokens (dotted line: $1$); it measures where the divergence
  concentrates, not the size of the gradient. (a) Over training, nine \opd{} runs with a complete per-role log
  (100 questions: four seeds; 300: seeds 1--3; 900: seeds 1--2), 10-step windows, mean and $\pm 1$ s.d. over runs.
  (b) One model per condition (zero-shot student; 300-question seed-1 \opd{} student) on 100 test questions,
  11{,}625 and 12{,}842 tokens (Appendix~\ref{app:heatmaps}).}
  \label{fig:mechanism}
\end{figure}

Figure~\ref{fig:mechanism} shows where the per-role feedback lands (the full training curves for both tasks
are in Appendix~\ref{app:geo-dynamics}). On the zero-shot student's outputs the answer line, 6\% of tokens,
holds 14\% of the KL (Appendix~\ref{app:heatmaps}); over the first ten \opd{} steps
on \chartqa{} it is $1.3\times$, by steps 11--20 $0.42\times$, and at the end $0.27\times$, while numeric-value
tokens stay at $0.29\times$ to $0.43\times$ their share and the mean per-token KL falls from $0.41$ to $0.15$:
the feedback is spent first on how the student ends its solution and is diffuse afterwards. The log measures
agreement, not correctness: a digit is not a perceptual token, and a low divergence means agreement whether
both models are right or wrong, so it does not show whether the student's reading of the chart improves.

\subsection{With an unreliable teacher, the student does not complete its solutions}
\label{sec:geo}

\begin{table}[t]
  \centering
  \scriptsize
  \setlength{\tabcolsep}{3pt}
  \caption{\geo: accuracy on 500 held-out problems, four seeds at every budget (means; across-seed sd between
  $0.006$ and $0.025$, per seed in Appendix~\ref{app:seeds}; every mean has a question-clustered 95\%
  interval of $\pm 0.035$ to $\pm 0.038$), the paired \opd{} minus \sft{} difference with its clustered and
  seed-level $t$ intervals, ``Completion'', the fraction of responses with an answer line (every response
  without one ran to the 1{,}536-token budget; Appendix~\ref{app:completion}), and ``Acc.\,$|$\,compl.'',
  accuracy among those (range over seeds). Bold as in Table~\ref{tab:chartqa}. Zero-shot student $0.366$
  (completion $0.65$), teacher $0.570$ (completion $0.78$).}
  \label{tab:geo}
  \begin{tabular}{lcccccccc}
    \toprule
    & \multicolumn{2}{c}{Accuracy} & \multicolumn{2}{c}{\opd{} $-$ \sft{}} & \multicolumn{2}{c}{Completion} & \multicolumn{2}{c}{Acc.\,$|$\,compl.} \\
    \cmidrule(lr){2-3} \cmidrule(lr){4-5} \cmidrule(lr){6-7} \cmidrule(lr){8-9}
    Problems & \sft & \opd & paired, clustered 95\% & $t$ over seeds & \sft & \opd & \sft & \opd \\
    \midrule
    100 & $0.423$ & $0.404$ & $-0.019$ [$-0.045$, $+0.007$] & [$-0.059$, $+0.021$] & $0.78$ & $0.64$ & $0.52$--$0.57$ & $0.61$--$0.69$ \\
    300 & $0.420$ & $0.419$ & $-0.001$ [$-0.024$, $+0.022$] & [$-0.020$, $+0.019$] & $0.73$ & $0.62$ & $0.55$--$0.59$ & $0.65$--$0.70$ \\
    900 & \tb{$0.481$} & $0.402$ & \tb{$-0.079$ [$-0.105$, $-0.053$]} & \tb{[$-0.110$, $-0.048$]} & $0.76$ & $0.63$ & $0.62$--$0.67$ & $0.63$--$0.65$ \\
    2101 & \tb{$0.478$} & $0.411$ & \tb{$-0.067$ [$-0.092$, $-0.043$]} & \tb{[$-0.094$, $-0.040$]} & $0.76$ & $0.64$ & $0.62$--$0.64$ & $0.61$--$0.66$ \\
    \bottomrule
  \end{tabular}
\end{table}

\begin{figure}[t]
  \centering
  \includegraphics[width=0.9\linewidth]{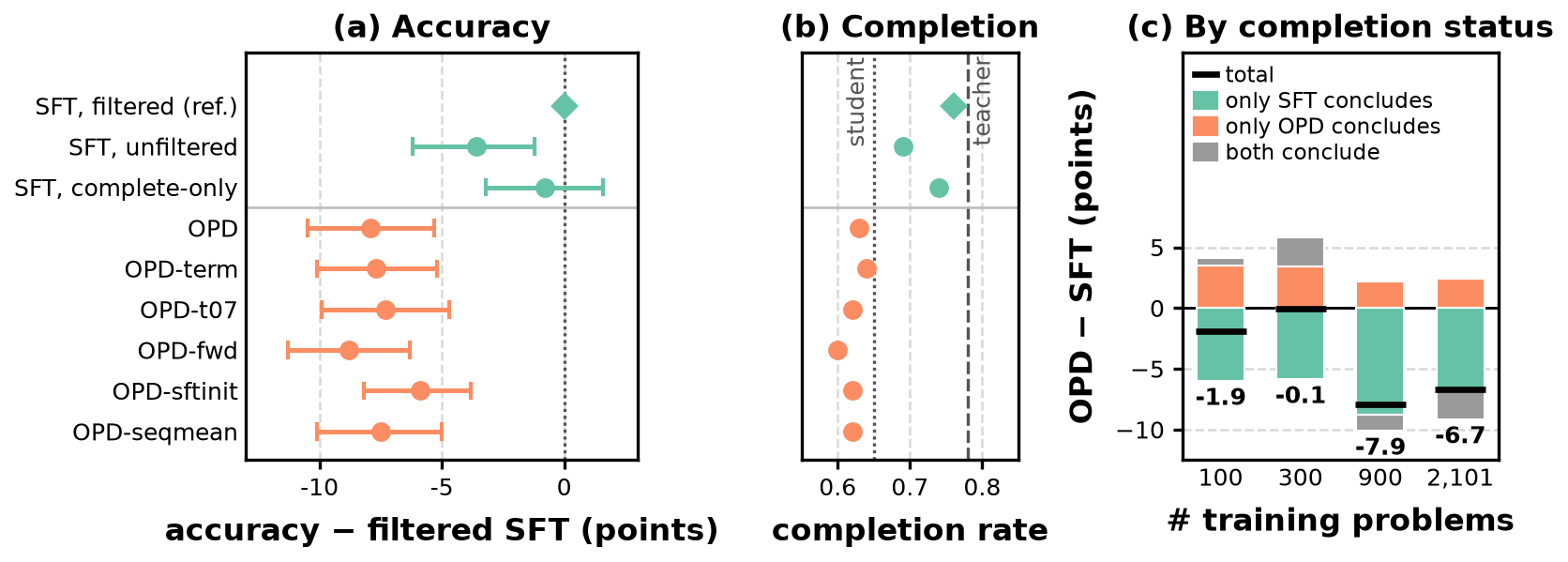}
  \caption{\geo, 900 problems, four seeds per row. (a) Paired accuracy difference of the three \sft{} data sets
  (filtered, the reference; unfiltered; complete-only, defined in the text) and of plain \opd{} and its five
  variants against the filtered \sft; bars: 95\% question-clustered bootstrap intervals with the four models per
  method as fixed (seed-level $t$ intervals in Table~\ref{tab:control}, Appendix~\ref{app:control300}).
  (b) Share of test responses with an answer line; dashed: zero-shot teacher; dotted: zero-shot student.
  (c) \opd{} minus \sft{} accuracy at each budget, in points of the 500 test problems, split by which method
  concluded the problem (a problem only one method concludes counts against the other); black tick: total.
  From the per-problem records; printed totals rounded to one decimal (Table~\ref{tab:decomp}).}
  \label{fig:completion}
\end{figure}

On \geo{} the ordering reverses (Table~\ref{tab:geo}, Figure~\ref{fig:curves}b). \sft{} leads by $1.9$
points at 100 problems (intervals include zero), the two methods tie at 300, and \sft{} leads by $7.9$ points
at 900 and $6.7$ at 2{,}101, every per-seed difference between $-5.2$ and $-10.6$ and both intervals excluding
zero. An \opd{} student concludes 62--64\% of test problems against 73--78\% for \sft, a completion gap of 11 to 14 points
at every budget, and every missing answer is a response that ran to the token budget (186 of 186 at 900
problems; Appendix~\ref{app:completion}). Among responses that do conclude, the
\opd{} student's accuracy is higher ($0.61$--$0.70$ against $0.52$--$0.67$), but the two methods conclude
different problems.
Decomposing the paired difference by which method concluded a problem (Figure~\ref{fig:completion}c;
Appendix~\ref{app:completion}; every
contribution in points of the 500-problem accuracy): at 900 problems the 95 problems per 500 that only \sft{}
concludes cost \opd{} $8.8$ points, the 29 that only \opd{} concludes gain it $2.2$, and the 285 both conclude
cost it $1.4$, so completion accounts for $6.6$ of the $7.9$ points; at 2{,}101 it accounts for $4.3$ of $6.7$.
Within the subset both methods conclude, \opd's accuracy is $2.4$ points below \sft's at 900 [$-6.0$, $+1.2$],
$4.0$ below at 2{,}101 [$-7.7$, $-0.3$], and $4.3$ above at 300 [$+2.7$, $+9.7$]: the higher conditional
accuracy of \opd{} in Table~\ref{tab:geo} is largely a selection effect, and the completion gap explains most,
not all, of the accuracy gap.

\paragraph{Is the teacher the source?}
Two observations point at the teacher. First, the
teacher's own solutions do not conclude: 78\% of its training samples and 78\% of its greedy test solutions
contain an answer line. Second, the student's failures concentrate on the problems the teacher does not finish
(Appendix~\ref{app:completion}): on the 109 test problems the greedy teacher leaves incomplete, the zero-shot,
\sft, and \opd{} students conclude 35\%, 45\%, and 20\%; on the 391 it concludes, 73\%, 85\%, and 75\%.
Problem difficulty could drive both failures, so this is an association. The most direct evidence is what the teacher does when it takes over a
student's state (Table~\ref{tab:continue}, Appendix~\ref{app:completion}): continued greedily from a 256-token prefix of an \opd{} student's
truncated test output with the rest of the budget, it concludes 34\%, and 3\% from the truncation point,
against 90\% from a 256-token prefix of a concluded output. That budget shrinks with the cut (1{,}280,
1{,}024, 512, and 512 tokens), so the probe was repeated with 512 tokens at every cut (lower block): the
within-trajectory difference between the 256-token cut and the truncation point falls from $+30.6$ [$+25.4$,
$+35.8$] to $+12.2$ [$+8.5$, $+16.2$] points for the \opd{} student and to $+0.8$ [$-5.1$, $+7.4$] for the
\sft{} student. At equal budget the teacher recovers few failed prefixes of either student (14\% of \sft's,
16\% of \opd's from 256 tokens) and most prefixes of concluded outputs (77 to 81\%), but those are different
problems and trajectories, so the probe describes the states rather than identifies what produced them. The
teacher-forced probe agrees: along outputs that ran to the budget the teacher's end-of-turn probability stays
below $10^{-3}$ at every position, against $0.98$ at the position where a concluded output stops, and its mass
on answer-opening tokens (a proxy) peaks at $0.12$ to $0.13$, against $0.63$ to $0.82$ where a concluded output's
answer line begins (Appendix~\ref{app:completion}).
Changing the teacher's prompt does not supply what is missing (Table~\ref{tab:teacherint}, on the zero-shot
student's failed prefixes from 300 training problems rather than the trained students' test outputs):
completion guidance leaves its completion at 19\% against 18\% ($+1.1$ [$-3.3$, $+5.5$] points), and the gold
answer lowers it to 7\% ($-11.0$ [$-18.7$, $-4.4$]), so no tested prompt establishes the improvement a
training-level intervention would need.

\paragraph{Does a larger generation budget close the gap?}
It does not. Re-scoring the same four 900-problem models per method with 4{,}096 generated tokens instead of
1{,}536 (Appendix~\ref{app:budget}), the \opd{} students' incomplete responses fall only from 186 to 168 of 500
(accuracy $+0.8$ [$-0.4$, $+1.9$]), whereas \sft's fall from 120 to 50 (accuracy $+4.0$
[$+2.5$, $+5.5$]), so the gap widens from $7.6$ to $10.8$ points [$-13.5$, $-8.2$].

\paragraph{Which part of the filter matters: three \sft{} data sets.}
We retrained \sft{} on the teacher's first sample for every problem, correct or not, concluded or not
(Figure~\ref{fig:completion}a,b; Table~\ref{tab:control} in Appendix~\ref{app:control300}). At 300 problems, where filtered \sft{} and \opd{} tie, removing the filter drops
\sft's completion rate to \opd's $0.62$ and its accuracy $1.8$ points below both (intervals include zero). At 900 problems the unfiltered \sft{} is $3.6$ points below the filtered one (per seed $-1.2$ to $-8.6$, so the seed-level interval includes zero), its completion rate ($0.69$) lies between
\sft's and \opd's, and it stays $4.2$ points above \opd{} [$+1.8$, $+6.7$]. Removing the filter changes the demonstrations' correctness, completeness, length, difficulty mix, and number at
once. A third \sft, trained on the teacher's first sample whenever it contains an answer line, correct or not
(1{,}645 of 2{,}101 samples; the criterion is the answer line, not an emitted end-of-turn token), separates
these: at 900 problems it reaches completion $0.74$ and accuracy $0.473$, within $2.1$ [$-4.4$, $+0.2$] and
$0.8$ [$-3.2$, $+1.6$] points of the correctness-filtered \sft{} and $4.7$ [$+2.5$, $+6.9$] and $2.9$ [$+0.7$,
$+5.0$] points above the unfiltered one (Table~\ref{tab:control}; Appendix~\ref{app:intervals}); at 300 problems it matches the filtered \sft{} ($0.72$ / $0.432$ against $0.73$
/ $0.420$). Selecting completed demonstrations thus recovers much of the observed \sft{} advantage under the
tested recipes. The difference between the two filtering schemes (correctness-filtered minus
complete-only) is $+0.8$ points [$-1.6$, $+3.2$], so a benefit of correctness of up to about three points is
not excluded, and the two data sets differ in size (1{,}163 against 1{,}645 demonstrations) as well as in
correctness, so this is a difference between recipes rather than an identified contribution of correctness; a
count-matched control was not run.

\paragraph{Three on-policy variants, and what their failure does and does not say about the objective.}
An on-policy analogue of the filter computes the loss only on rollouts that terminated (\opd-term); rollouts at the teacher's temperature test the sampling distribution
(\opd-t07); the forward KL tests whether the reverse KL attenuates the stopping signal (\opd-fwd). None of the
three restores completion (Figure~\ref{fig:completion}a,b): it stays at $0.60$ to $0.64$, accuracy within a point of plain \opd,
and all three match plain \opd{} on \chartqa{} (Appendix~\ref{app:chartqa-variants}). The gradient shows why
the divergence alone cannot be blamed: the loss compares the two next-token distributions over the whole
vocabulary at every position, so the teacher's probability of stopping enters the loss whether or not the
student sampled a stop. Writing $p_e$ and $q_e$ for the student's and the teacher's
end-of-turn probabilities at a position, $z_e$ for the student's end-of-turn logit, and $L$ for the position's
reverse KL, $\partial L / \partial z_e = p_e\,(\log(p_e/q_e) - L)$, which pushes $z_e$ up whenever $\log(p_e/q_e) < L$, that is $p_e < q_e e^{L}$ (a student
$p_e$ below $q_e$ suffices but is not necessary), but is scaled by $p_e$, so a
student that rarely stops receives a small push; the forward KL has gradient
$p_e - q_e$, without the attenuation. That \opd-fwd fails as plain \opd{}
does ($0.394$, completion $0.60$) argues against attenuation and leaves two candidates: the teacher, conditioned
on the student's overlong prefixes, may itself assign little probability to stopping; and the token-mean loss of
Equation~\ref{eq:opd} is dominated by rollouts that ran to the budget, which hold about three quarters of the
positions in the 900-problem runs (mean rollout length 892 tokens, termination rate $0.55$). The second does not by
itself explain the failure: a sequence-mean loss, in which every rollout counts once (\opd-seqmean;
Appendix~\ref{app:hparams}), does not restore completion, with $0.407$ accuracy and $0.62$ completion at 900 problems
against $0.402$ / $0.63$ for plain \opd{} ($+0.4$ [$-1.4$, $+2.1$] and $-0.7$ [$-2.6$, $+1.2$]; per seed in
Appendix~\ref{app:intervals}), the same termination trajectory during training, and $0.825$ against $0.826$ on
\chartqa{} (Table~\ref{tab:control}, Appendix~\ref{app:chartqa-variants}). The first is what the probe above
measures: $q_e$ is below $10^{-3}$ on these prefixes, so the target the objective sets at these positions is a
student that does not stop either. Whether the gradient at a position pushes $z_e$ up depends on the student's $p_e$
and on the position's full KL, neither of which we recorded, so the direction of the push is not measured:
the probe establishes the target, not the gradient. Filtered \sft{} instead concentrates its supervision on
demonstrations that contain an answer line.

\paragraph{On-policy feedback undoes the completion gain the student obtained from \sft.}

We started \opd{} from the merged filtered-\sft{} student of the same seed (\opd-sftinit), which concludes 76\%
of its test solutions, and trained it for the usual 150 steps. The training log
(Figure~\ref{fig:curves}d) shows the behaviour being undone: the rollout termination rate is $0.79$ over the first
five steps, $0.54$ by steps 21--40, and $0.60$ over the rest of the run,
against $0.63$, $0.48$, and $0.56$ for plain \opd{} from the base model. At evaluation it concludes 62\% of test
responses, indistinguishable from plain \opd{} ($-0.5$ [$-2.8$, $+1.7$]) and $13.8$ points below the \sft{} model it
started from ([$-16.4$, $-11.2$]; $t$ [$-17.7$, $-9.8$]). On the problems both it and \sft{} conclude no difference is
detected ($-0.5$ [$-3.6$, $+3.1$]); it ends at $0.422$, $5.9$ points below \sft{} ([$-8.2$, $-3.8$]; $t$ [$-10.9$,
$-1.0$]) and $1.9$ above plain \opd{} ([$-0.2$, $+4.2$], not robust). The objective thus undoes a completion gain the student already had rather than merely failing to teach one; the
experiment identifies the training process as the cause, not which part of it.

\subsection{Distribution shift and teacher size}
\label{sec:ood}

Off distribution (Appendix~\ref{app:ood}), both methods move the student
by only 2 to 5 points on CharXiv and ChartQAPro, against a teacher that is itself at $0.41$ there; the only
paired difference that excludes zero is \opd's $+3.0$ [$+0.5$, $+5.4$] on CharXiv at 100 questions, the other
seven are within $\pm 0.7$ points, and overlapping intervals do not establish equivalence.
A 32B teacher on the 300-question \chartqa{} run (four seeds) is itself $1.4$ points more accurate ($0.858$
against $0.844$, an interval that includes zero) and its student is not more accurate ($0.812$ against
$0.826$, $-1.5$ [$-3.1$, $+0.2$]; $+1.2$ [$-0.9$, $+3.4$] against \sft).

\section{Related work}

\paragraph{On-policy distillation.}
Building on imitation-learning distillation \citep{lin2020imitkd}, generalised knowledge distillation
\citep{agarwal2024gkd}, MiniLLM \citep{gu2024minillm}, and DistiLLM \citep{ko2024distillm} established
student-sampled, teacher-scored token-level distillation, and \citet{agarwal2024gkd} reported the
data-efficiency advantage that Section~\ref{sec:chartqa} measures for a VLM; text-only analyses \citep{li2026rethinking,xie2026position}
study its phenomenology. Concurrent text-only work argues that teacher reliability must be verified before dense on-policy supervision
\citep{zhang2026tgopd}, that token-level corrections cannot unfold a complete repair \citep{sun2026sopd}, and
that the teacher's supervision is noisy \citep{ding2026reallydistill};
\citet{qiu2026opdaha} show for visual hallucination that when teacher and student condition on the same flawed
prefix the teacher's signal collapses toward the student's error. \citet{ma2026outcome} shows that low
teacher--student token-level divergence can coexist with an incorrect final answer in text-only \opd. We study
a distinct failure mode in vision-language distillation: responses that never reach an answer line. We
decompose the resulting accuracy gap by completion status and show that five on-policy variants fail to restore
completion; \opd{} also erases the completion gain of an \sft-initialised student. Changing the divergence
\citep[see also][]{xiao2026tvopd} or weighting rollouts \citep[as][propose offline]{ai2026learnability} did not
restore completion.

\paragraph{On-policy distillation for vision-language models.}
A rapid 2026 literature adapts the objective to VLMs: visual-advantage weighting \citep{liu2026vaopd} up-weights
the tokens whose teacher prediction depends on fine visual detail and, trained on \geo{} and ViRL39K, reports
gains over plain \opd{}; decomposed gradient
steering \citep{yoon2026decomposed} and Fisher-projected distillation \citep{xue2026fpopd} reshape the update
toward grounding; others transfer reasoning from a text-only teacher \citep{bousselham2026vold} or replace the
teacher with the same model given privileged input
\citep{yuan2026visionopd,zhu2026rpopsd,xiang2026ptdpo,wang2026videoopsd}. All change the objective and
compare the change against plain \opd; we keep the objective fixed and ask how it compares with filtered
demonstrations and which behaviour of the teacher the student acquires.

\section{Conclusion}
We study when OPD improves over filtered SFT and how teacher behavior on student-generated trajectories affects the student.
OPD is data-efficient when the teacher reliably completes its solutions, but degrades when the teacher frequently fails to complete.
On Geometry3K, most of the gap comes from incomplete responses, and OPD even erases completion gains previously learned through SFT.
Filtering teacher demonstrations for completeness alone recovers much of SFT's advantage, while teacher probes show that the teacher itself rarely completes from failed student trajectories.
These results highlight teacher reliability on student-visited states as a critical factor in the success of OPD.

\subsection*{Limitations}
Our evaluation covers one model family, two tasks, one student size below 4B, and a second teacher size on
one task; other families, tasks, and scales remain untested. The
attribution of the completion failure to the teacher rests on the teacher's own trajectories, the per-problem
overlap with the student's failures, and the probe on the student's prefixes; no tested teacher
prompt clearly improved its recovery on the student's failed prefixes (one lowered it), so a teacher trained to
conclude from these states remains untested; we did not record the student's own end-of-turn probability on those prefixes, and the
continuation probe compares truncated and concluded outputs on different problems. With four seeds per point, several secondary differences depend on how the interval is formed
(Appendix~\ref{app:intervals}), and the \sft{} training length was selected on the test set. The unfiltered control removes both filters and its data carry an appended end-of-turn token; the
completeness-only control separates the filters, not the token, and has more demonstrations. Truncation is
inferred by re-tokenising (validated in Appendix~\ref{app:completion}); token roles are heuristic.

\subsection*{AI use statement}
We used AI coding assistants (Claude, Anthropic) to assist in this project. It assisted with writing and refactoring training, evaluation, and analysis code under the authors' direction, drafted the documentation and the
prose of this manuscript, which the authors revised, and assisted with literature search. Every experiment was
designed by the authors and run by them on their own compute; every number in the paper is read from the
released per-question evaluation records, and the statistical procedures were checked against independent
implementations in the test suite. The authors reviewed all generated code and text
and take full responsibility for the content.

\subsection*{Ethics statement}
We strictly adhere to the ICLR Code of Ethics. This work distills publicly released models on public benchmarks and introduces no new data collection. The
distilled students inherit the biases and failure modes of the teacher, which is one of the paper's findings;
we do not release them as general-purpose assistants.

\subsection*{Reproducibility statement}
All code, configurations, Slurm scripts, per-step training logs, and per-question evaluation records
(prediction, gold answer, correctness, and completion for every test question of every model) are released at
\url{https://github.com/ffy208/opd-inherits-teacher-vlm}; every mean, interval, and paired
difference in the paper is recomputed from the records by one script. Data order, rollout sampling, and LoRA
initialisation are seeded, so a rerun follows the same recipe, but GPU kernels and vLLM decoding are not
bit-reproducible across hardware, and re-evaluating the same weights can move a single seed's accuracy by a few
tenths of a point on \chartqa{} and by up to $2.8$ points on \geo{} (Section~\ref{sec:setup}). The released
evaluator records each generation's stop reason and token count, which the records used here lack
(Appendix~\ref{app:completion}). Appendix~\ref{app:hparams} lists every hyperparameter, the test-subset
construction, the answer parser, and the end-of-turn convention; Appendix~\ref{app:engineering} lists the
engineering choices needed to train a VLM student and teacher on one 40 GB GPU.

\subsection*{Acknowledgments}
We thank Geyang Guo for her detailed feedback and helpful suggestions for improving the manuscript. We acknowledge computational resources and services provided by the AI Makerspace of the College of Engineering
(RRID:SCR\_028058) and the Partnership for an Advanced Computing
Environment (PACE) at the Georgia Institute of Technology,
Atlanta, Georgia, USA (RRID:SCR\_027619).

\bibliography{refs}
\bibliographystyle{iclr2027_conference}

\clearpage
\appendix
\raggedbottom

\section{Hyperparameters and evaluation details}
\label{app:hparams}

Table~\ref{tab:hparams} lists every training, sampling, and evaluation setting; the end-of-turn convention and
the answer parser are the ones the loss mask and the scorer use.

\begin{table}[H]
  \centering
  \small
  \caption{Hyperparameters, data construction, and evaluation settings.}
  \label{tab:hparams}
  \begin{tabular}{p{2.3cm}p{10.4cm}}
    \toprule
    Student / teacher & Qwen3-VL-2B-Instruct / Qwen3-VL-8B-Instruct (bf16); Qwen3-VL-32B-Instruct in Section~\ref{sec:ood} \\
    LoRA & rank 64, $\alpha = 128$, language-model q/k/v/o and MLP projections, vision encoder and projector frozen \\
    Data & 500 test questions sampled with seed 42 from the official test split (human-written \chartqa{} questions; the 601 \geo{} test problems); training budgets are nested prefixes of a seeded shuffle of the training split; images resized to a longest side of 768 px (300 to 600 visual tokens) \\
    \sft & 302 steps, batch 4 $\times$ accumulation 4, lr $10^{-4}$, cosine, max sequence length 2{,}048; loss on the response tokens only; an end-of-turn token is appended to every demonstration before the sequence is cut to 2{,}048 tokens (the prompt is never cut) \\
    \sft{} data & one teacher sample per question at $T{=}0.7$, kept if its answer is correct (2{,}402 of 3{,}000 on \chartqa; 1{,}163 of 2{,}101 on \geo, of whose samples 78\% contain an answer line); the unfiltered control keeps every first sample; the complete-only control keeps the first sample when it contains an answer line, correct or not (1{,}645 of 2{,}101) \\
    \opd & 150 steps, 16 rollouts per step, micro-batch 4, lr $5\cdot10^{-5}$, 20 warmup steps, cosine, AdamW \citep{loshchilov2017adamw} without weight decay, grad clip 1.0 \\
    Rollouts & \texttt{<|im\_end|>} is the end-of-turn token; a rollout ends at its first occurrence or at the budget of 512 new tokens (charts) / 1{,}536 (geometry); $T{=}1.0$, top-$k$ off \\
    Divergence & reverse KL over the full vocabulary (end-of-turn token included) in fp32 chunks, at every generated position up to and including the end-of-turn token; loss is the mean over all such positions of the batch (Equation~\ref{eq:opd}); teacher and student score the same prefixes with the same attention mask \\
    Variants & one change each: \opd-term (loss on rollouts that emitted end-of-turn only), \opd-t07 (rollouts at $T{=}0.7$), \opd-fwd (forward KL), \opd-sftinit (student initialised from the merged \sft{} model of the same seed), \opd-seqmean (sequence-mean loss $\frac{1}{B}\sum_{i}\frac{1}{|y^{(i)}|}\sum_{t} D^{(i)}_t$ over the $B{=}16$ rollouts of a step, so that every rollout counts once instead of once per token, in place of the token mean of Equation~\ref{eq:opd}), \sft{} complete-only (the data above), \sft-$n$ ($n$ steps instead of 302) \\
    Evaluation & vLLM greedy decoding, the task's token budget (4{,}096 in Appendix~\ref{app:budget}); the answer is the last line matching \texttt{Answer:}; relaxed accuracy with 5\% relative numeric tolerance (exact match for a gold value of 0; normalised string match for text answers); radicals, fractions, and $\pi$ are evaluated numerically \\
    Compute & one A100 40 GB or L40S 48 GB per point; \opd{} 13.5 s/step (charts), 85 s/step (geometry) \\
    \bottomrule
  \end{tabular}
\end{table}

\paragraph{Prompts.}
Every model receives one image and the following user message (the \chartqa{} version replaces the first two
sentences by ``Look at the chart and answer the question below. Reason step by step: first read the relevant
values from the chart, then do any calculation needed.''); the student always sees the geometry version.
\begin{quote}\small\ttfamily
Look at the geometry figure and solve the problem below.\\
Reason step by step: first state the given measures and relations you can read from the figure, then apply the
relevant theorems and compute. Give the final answer as a number (decimals are fine; simplify radicals or
fractions to a decimal if needed). [\textrm{\emph{guidance or paraphrase inserted here}}]Keep the reasoning
concise.\\
On the last line, write the final answer in exactly this format:\\
Answer: <your answer>\\
\\
Question: \{question\}[\textrm{\emph{privileged suffix appended here}}]
\end{quote}
The completion guidance inserted before ``Keep the reasoning concise'' is: ``Do not repeat a step you have
already completed or restart the solution; once the value is determined, stop reasoning and commit to it, and
always finish with the answer line.'' The length-matched paraphrase replaces the first two sentences by ``Study
the geometry figure and work out the problem below. Proceed step by step: begin with the measures and
relationships shown in the figure, then use the appropriate theorems to compute the result. Report the final
answer as a number (decimals are acceptable; convert radicals or fractions to a decimal when needed).'' and
inserts ``Name each quantity you use as you introduce it, so that the intermediate values of the solution can
be read directly from the reasoning, and give the answer in the units of the figure.'' at the same place. The
privileged suffix is ``The correct final answer is known to be: \{answer\}. Write out the reasoning that leads
to it, and finish with the same `Answer: <your answer>' line.'' In the continuation probe the prompt is followed
by the assistant turn containing the student's prefix, which the teacher continues greedily.

\section{Feedback dynamics}
\label{app:geo-dynamics}

\begin{figure}[H]
  \centering
  \includegraphics[width=\linewidth]{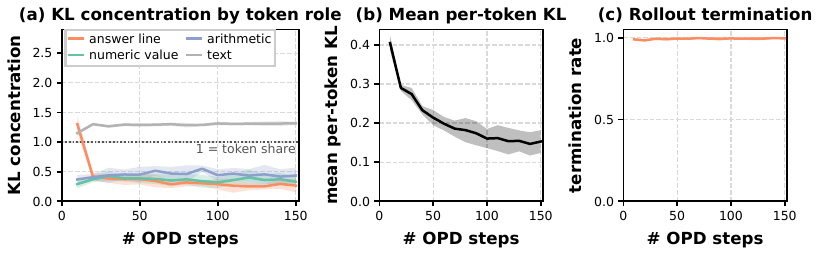}
  \caption{Feedback dynamics on \chartqa{} over the nine \opd{} runs with a complete per-role training log
  (100 questions: four seeds; 300: seeds 1--3; 900: seeds 1--2; the 300-question seed-42 run predates per-role
  logging and the other logs stop at an earlier checkpoint on the Hub; windows of 10 steps, shaded: $\pm 1$
  s.d. over runs). (a) KL concentration per token role (share of the batch's KL divided by share of tokens;
  dotted line: $1$); (b) mean per-token KL; (c) rollout termination rate.}
  \label{fig:dynamics}
\end{figure}

\begin{figure}[H]
  \centering
  \includegraphics[width=\linewidth]{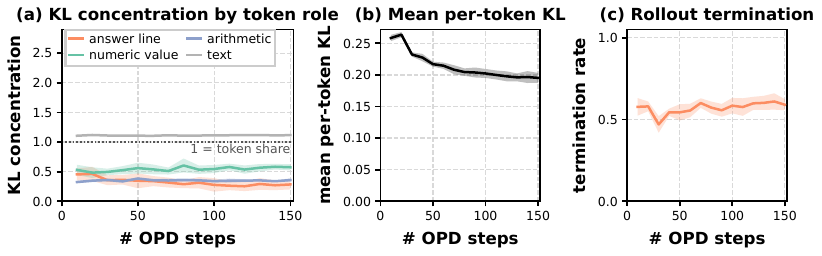}
  \caption{As Figure~\ref{fig:dynamics}, for the eleven \opd{} runs on \geo{} with a complete per-role log
  (300 and 900 problems: four seeds each; 100 problems: seeds 1 and 42; 2{,}101 problems: seed 42; the Hub
  copies of the other 100- and 2{,}101-problem logs stop at an earlier checkpoint). The answer-line
  concentration falls from $0.45$ to $0.28$; numeric values stay between $0.48$ and $0.60$; the rollout
  termination rate (c) stays between $0.47$ and $0.61$ over 150 steps.}
  \label{fig:geodyn}
\end{figure}

Figure~\ref{fig:dynamics} gives the per-role feedback log of Section~\ref{sec:dynamics} on \chartqa{} and
Figure~\ref{fig:geodyn} repeats the analysis for the \geo{} runs. We also tested
whether the end-of-training KL share on numeric tokens predicts a run's remaining gap to the teacher across runs:
with 9 runs on \chartqa{} and 11 on \geo{} the Spearman correlations are $-0.13$ [$-0.83$, $+0.73$] and $+0.66$
[$+0.07$, $+0.89$]; the \geo{} interval excludes zero narrowly, but it is one of five statistics we examined and
the runs' accuracies within a task are too close for this test to have power, so we do not claim it.

\section{Token-level feedback before and after \opd}
\label{app:heatmaps}

\begin{figure}[H]
  \centering
  \includegraphics[width=\linewidth]{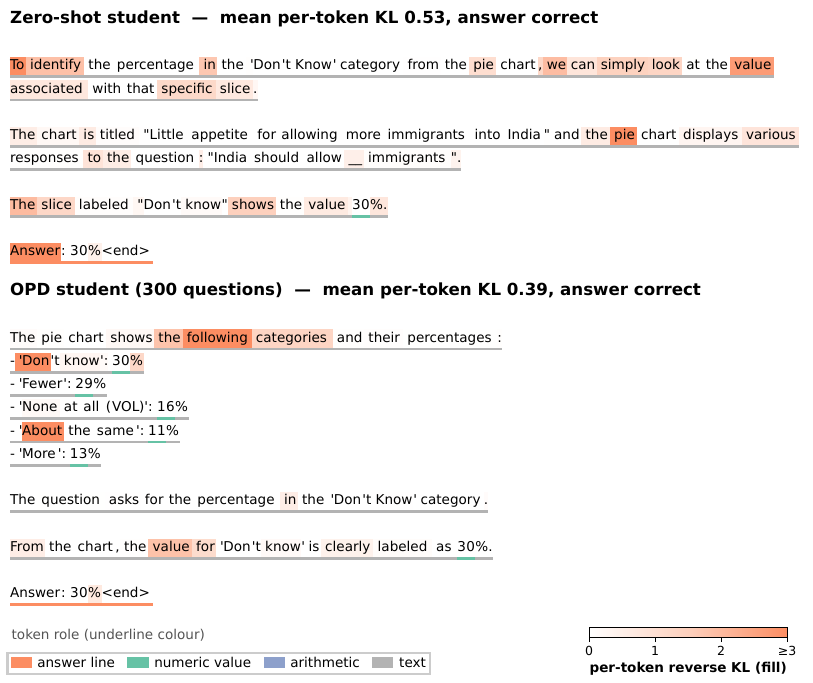}
  \caption{Per-token reverse KL between teacher and student on one \chartqa{} test question (rollouts at
  temperature 1.0), for the zero-shot student (top) and the \opd{} student trained on 300 questions (bottom),
  same colour scale; the underline gives the token's role. Before training the opening of the answer line
  carries a large divergence; after training it carries almost none and the residual disagreement is diffuse
  (here largest on a wording choice). Numbers are tokenised digit by digit and the disagreement on a number sits
  on its first digit.}
  \label{fig:heatmaps}
\end{figure}

\begin{figure}[H]
  \centering
  \includegraphics[width=\linewidth]{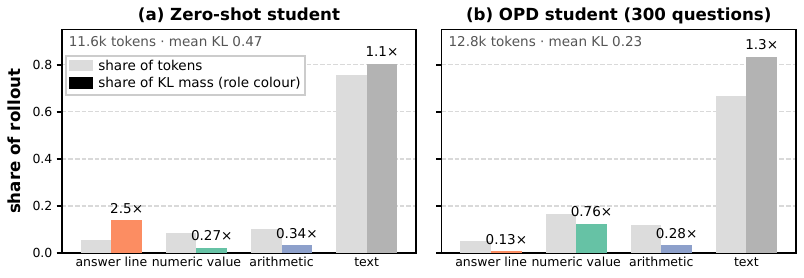}
  \caption{Share of tokens against share of the teacher's KL mass per role over the rollouts on 100 \chartqa{}
  test questions (11{,}625 and 12{,}842 tokens), for the zero-shot student (a) and the \opd-300 student (b,
  seed 1); the number above each bar is the concentration (KL share divided by token share). Mean per-token KL
  falls from $0.47$ to $0.23$.}
  \label{fig:tokenclasses}
\end{figure}

Figure~\ref{fig:heatmaps} shows the per-token divergence on one test question before and after training, and
Figure~\ref{fig:tokenclasses} the role breakdown over 100 test questions that the concentration statistic of
Section~\ref{sec:setup} summarises.

\section{Completion, termination, and truncation}
\label{app:completion}

\paragraph{How truncation is identified.}
The evaluation records used in this paper store each model's output text but not the generator's stop reason.
Truncation is therefore inferred: the output is re-tokenised with the student's tokenizer and counted as
truncated if it lies within 8 tokens of the generation budget (the re-tokenised lengths of truncated outputs
cluster at the budget minus 0 to 2 tokens; only one output in the 900-problem \opd{} runs falls between 8 and 40
tokens below it). ``Stops'' in Table~\ref{tab:crosstab} means not truncated by this criterion; it is an
approximation of an emitted end-of-turn token, not a direct observation of one. The released evaluator now
records the stop reason and output token count of every generation, so future records need no inference.
Re-evaluating the 900-problem \opd{} and \sft{} students of seed 1 with that evaluator (models rebuilt from
the Hub checkpoint and adapter; accuracy $0.422$ and $0.486$ against $0.406$ and $0.472$ in the recorded pass),
the criterion agrees with the recorded stop reason on 500 of 500 and 498 of 500 responses: it calls 183 and 120
responses truncated, the generator 183 and 118; the two disagreements are \sft{} responses that stopped within 8
tokens of the budget, and no truncated response is missed.

\paragraph{Teacher probe.}
Table~\ref{tab:probe} gives the teacher-forced probe: a pass of the 8B teacher over every stored test output of
a student (900-problem models, four seeds each; the zero-shot student is one model), recording at every prefix
length its probability of the end-of-turn token and of a token that opens an answer line: the three tokens
\texttt{Answer}, \texttt{Answer} with a leading space, and \texttt{**}, which are the first tokens of
\texttt{Answer}, \texttt{ Answer}, \texttt{**Answer}, and \texttt{Answer:} and the token after the newline in
\texttt{\textbackslash nAnswer}. This mass is a proxy for opening an answer line, not the probability of
one: \texttt{**} also opens ordinary bold text, and an answer line needs the following tokens as well. The
probabilities are computed in bf16 and summarised (maximum, mean, value at
the stop position) at full precision; only the stored per-position series are rounded to four decimals. ``Max
EOS'' in Table~\ref{tab:probe} is the maximum over the positions of one output, averaged over outputs; the
largest value at any position of any truncated output is $4 \cdot 10^{-5}$ (zero-shot), $8 \cdot 10^{-6}$
(\sft), $5 \cdot 10^{-5}$ (\opd), and $6 \cdot 10^{-4}$ (\opd-sftinit), so no truncated output has a position
above $10^{-3}$.

\begin{table}[H]
  \centering
  \scriptsize
  \setlength{\tabcolsep}{2pt}
  \caption{Continuation probe on \geo: share of the teacher's greedy continuations that reach an answer line,
  from prefixes of a student's truncated outputs (cut at 256, 512, and 1{,}024 tokens and at the truncation
  point) and, for comparison, from 256-token prefixes of its concluded outputs; the rest of the 1{,}536-token
  budget is 1{,}280, 1{,}024, 512, and 512 tokens. Seeds 1 and 42 of the 900-problem students pooled; $n$: truncated outputs in parentheses, and 203, 435, and
  320 concluded outputs of the zero-shot, \sft, and \opd{} students; brackets: 95\% question-clustered
  bootstrap intervals.}
  \label{tab:continue}
  \begin{tabular}{p{1.05cm}lccccc}
    \toprule
    & & \multicolumn{4}{c}{Truncated outputs, prefix cut at} & Concluded outputs \\
    \cmidrule(lr){3-6} \cmidrule(lr){7-7}
    Budget & Student ($n$) & 256 tokens & 512 tokens & 1{,}024 tokens & truncation point & 256 tokens \\
    \midrule
    \multirow{3}{=}{Rest of budget} & Zero-shot (176) & $0.35$ [$0.28$,\,$0.42$] & $0.13$ [$0.09$,\,$0.18$] & $0.01$ [$0.00$,\,$0.02$] & $0.01$ [$0.00$,\,$0.02$] & $0.91$ \\
    & \sft{} (241) & $0.32$ [$0.25$,\,$0.38$] & $0.24$ [$0.19$,\,$0.30$] & $0.07$ [$0.04$,\,$0.11$] & $0.11$ [$0.07$,\,$0.15$] & $0.87$ \\
    & \opd{} (369) & $0.34$ [$0.29$,\,$0.39$] & $0.17$ [$0.13$,\,$0.21$] & $0.06$ [$0.04$,\,$0.09$] & $0.03$ [$0.01$,\,$0.05$] & $0.90$ \\
    \midrule
    \multirow{3}{=}{Fixed 512 tokens} & Zero-shot (176) & $0.21$ [$0.15$,\,$0.27$] & $0.10$ [$0.06$,\,$0.14$] & $0.01$ [$0.00$,\,$0.02$] & $0.01$ [$0.00$,\,$0.02$] & $0.78$ \\
    & \sft{} (241) & $0.14$ [$0.10$,\,$0.19$] & $0.12$ [$0.08$,\,$0.16$] & $0.08$ [$0.04$,\,$0.12$] & $0.13$ [$0.09$,\,$0.18$] & $0.77$ \\
    & \opd{} (369) & $0.16$ [$0.12$,\,$0.20$] & $0.08$ [$0.06$,\,$0.11$] & $0.05$ [$0.03$,\,$0.08$] & $0.04$ [$0.02$,\,$0.06$] & $0.81$ \\
    \bottomrule
  \end{tabular}
\end{table}

Table~\ref{tab:continue} and Table~\ref{tab:continueapp} give the continuation
probe: the teacher continues each output greedily from a prefix of 256, 512, or 1{,}024 tokens with the rest of
the 1{,}536-token budget, at least 512 tokens (so 1{,}280, 1{,}024, and 512), and from the truncation point with
512 more tokens, and a second pass gives every cut the same 512 tokens (lower block of
Table~\ref{tab:continue}); prefix plus continuation is scored like a response. Per seed, the fixed-budget
completion rate from 256 tokens is $0.16$ / $0.16$ for the two \opd{} students and $0.12$ / $0.16$ for the two
\sft{} students. Prefixes that already contain an answer line
are not continued (nine prefixes of concluded outputs; none of a truncated output). The intervals of
Table~\ref{tab:continue} are question-clustered percentile bootstraps over the pooled seeds (the two outputs of
a problem are resampled together), and the paired differences between cuts keep each trajectory's pairing; per
seed, the
completion rate from the truncation point is $0.02$ and $0.04$ for the two \opd{} students and $0.08$ and $0.13$
for the two \sft{} students, and from 256 tokens $0.34$ / $0.33$ and $0.26$ / $0.37$.

\paragraph{Prompt-level interventions on the teacher.}
Table~\ref{tab:teacherint} repeats the fixed-budget continuation with the teacher under three other prompts, on
the outputs of the zero-shot student for the first 300 training problems (91 truncated, 209 concluded; the test
set is not used to design an intervention, and these are not the trained students' test outputs of
Table~\ref{tab:continue}): the geometry prompt with explicit completion guidance appended, a length-matched
paraphrase of the geometry prompt without that guidance, and the geometry prompt with the gold answer appended
(privileged information); the full texts are in Appendix~\ref{app:hparams}. The student's prompt is unchanged
in every case. Paired over the same prefixes, completion guidance minus the original prompt is $+1.1$ [$-3.3$,
$+5.5$] points of completion at the 256-token cut and $+2.2$ [$-2.2$, $+6.6$] at 512, and $-1.1$ [$-3.3$,
$+0.0$] and $+1.1$ [$+0.0$, $+3.3$] points of accuracy; guidance minus the paraphrase is $+1.1$ [$-7.7$, $+9.9$]
and $+2.2$ [$+0.0$, $+5.5$] of completion, $+1.1$ [$-2.2$, $+4.4$] and $+1.1$ [$+0.0$, $+3.3$] of accuracy: no
improvement is detected; at the 256-token cut the interval against the original prompt excludes improvements
above 5.5 points, and the less precise comparison with the paraphrase excludes only improvements above 9.9 points. The
gold answer lowers completion by $11.0$ points [$-18.7$, $-4.4$] at 256 tokens and from $0.78$ to $0.54$ on the
concluded outputs; 85 of the 91 continuations run to the budget (75 under the original prompt). Reading the
continuations, 37 of the 91 re-examine an earlier step (22 under the original prompt), 20 note a contradiction
between the prefix and what they derive (15), and 6 refer to the given answer; for example, one continuation
computes the altitude of a right triangle, then writes ``Wait, that's not correct. The altitude to the
hypotenuse is the geometric mean of the two segments only if the altitude is drawn to the hypotenuse'' and
restarts. Re-derivation is thus a plausible reading of the drop, not a demonstrated cause. Because no tested
prompt established the improvement in teacher recovery that the training-level intervention (the same students
trained under such a teacher) was to build on, that intervention was not run; it remains open whether a teacher
trained to conclude from these states transmits completion through on-policy distillation.

\begin{table}[H]
  \centering
  \footnotesize
  \caption{Fixed-budget continuation (512 tokens) by the teacher under four prompts, from prefixes of the
  zero-shot student's outputs on the first 300 training problems of \geo: completion / accuracy on the 91
  truncated outputs (brackets: 95\% bootstrap interval of the completion rate over outputs, one per problem) and
  on the outputs that concluded ($n$ in parentheses; concluded outputs shorter than a cut are not continued).
  Paired differences between prompts are in the text.}
    \label{tab:teacherint}
  \setlength{\tabcolsep}{3pt}
  \begin{tabular}{lccc}
    \toprule
    Teacher prompt & Truncated, 256 & Truncated, 512 & Concluded, 256 \\
    \midrule
    Original & $0.18$ [$0.10$, $0.25$] / $0.08$ & $0.03$ / $0.00$ & $0.78$ / $0.49$ (138) \\
    With completion guidance & $0.19$ [$0.11$, $0.26$] / $0.07$ & $0.05$ / $0.01$ & $0.80$ / $0.49$ (138) \\
    Length-matched paraphrase & $0.18$ [$0.10$, $0.25$] / $0.05$ & $0.03$ / $0.00$ & $0.81$ / $0.50$ (138) \\
    With the gold answer & $0.07$ [$0.02$, $0.12$] / $0.07$ & $0.01$ / $0.01$ & $0.54$ / $0.54$ (138) \\
    \bottomrule
  \end{tabular}
\end{table}

\begin{table}[H]
  \centering
  \footnotesize
  \caption{Teacher-forced probe on \geo{} test outputs (mean over seeds; $n$ per 500). For outputs that conclude:
  the teacher's end-of-turn probability at the position where the student stopped (``EOS at stop''), the share of
  outputs on which it exceeds $0.5$, and its probability mass on answer-opening tokens (a proxy, see text) where the student's answer line
  begins (``answer at onset''). For outputs that ran to the budget: the maximum end-of-turn probability over the
  positions of an output, averaged over outputs (no single position exceeds $10^{-3}$), and the mean over outputs
  of the maximum answer-opening probability (above $0.5$ on 10 to 12\% of outputs).}
  \label{tab:probe}
  \setlength{\tabcolsep}{4pt}
  \begin{tabular}{lcccccc}
    \toprule
    & \multicolumn{3}{c}{Concluded outputs} & \multicolumn{3}{c}{Truncated outputs} \\
    \cmidrule(lr){2-4} \cmidrule(lr){5-7}
    Student & $n$ & EOS at stop (share $> 0.5$) & answer at onset & $n$ & max EOS & max answer \\
    \midrule
    Zero-shot & 324 & $0.98$ ($0.99$) & $0.63$ & 176 & $< 0.001$ & $0.12$ \\
    \sft & 380 & $0.98$ ($0.99$) & $0.77$ & 120 & $< 0.001$ & $0.13$ \\
    \opd & 314 & $0.98$ ($0.98$) & $0.76$ & 186 & $< 0.001$ & $0.13$ \\
    \opd-sftinit & 311 & $0.98$ ($0.98$) & $0.82$ & 189 & $< 0.001$ & $0.12$ \\
    \bottomrule
  \end{tabular}
\end{table}

\begin{table}[H]
  \centering
  \footnotesize
  \setlength{\tabcolsep}{3pt}
  \caption{Continuation probe on \geo{} (900-problem \sft{} and \opd{} students, seeds 1 and 42 pooled; the
  zero-shot student): completion / accuracy of the teacher's continuation from a prefix of the student's output,
  split by whether the student's own output concluded, with the number of outputs continued at each cut
  (concluded outputs shorter than a cut, and prefixes that already contain an answer line, are not continued).
  The truncated rows are those of the upper block of Table~\ref{tab:continue}.}
  \label{tab:continueapp}
  \begin{tabular}{llcccc}
    \toprule
    Student & Its output ($n$ per cut) & 256 tokens & 512 tokens & 1{,}024 tokens & Truncation point \\
    \midrule
    Zero-shot & truncated (176) & $0.35$ / $0.19$ & $0.13$ / $0.09$ & $0.01$ / $0.01$ & $0.01$ / $0.01$ \\
              & concluded (203 / 45 / 4) & $0.91$ / $0.53$ & $0.80$ / $0.27$ & $0.50$ / $0.25$ & \\
    \sft & truncated (241) & $0.32$ / $0.18$ & $0.24$ / $0.15$ & $0.07$ / $0.05$ & $0.11$ / $0.08$ \\
         & concluded (435 / 163 / 63) & $0.87$ / $0.57$ & $0.70$ / $0.44$ & $0.62$ / $0.41$ & \\
    \opd & truncated (369) & $0.34$ / $0.21$ & $0.17$ / $0.10$ & $0.06$ / $0.04$ & $0.03$ / $0.02$ \\
         & concluded (320 / 108 / 23) & $0.90$ / $0.58$ & $0.79$ / $0.44$ & $0.57$ / $0.39$ & \\
    \bottomrule
  \end{tabular}
\end{table}

\paragraph{Cross-tabulation.}
Table~\ref{tab:crosstab} cross-tabulates, for every model, whether a test response contains an answer line and
whether it was truncated. For every correctness-filtered \sft{} and \opd{} model, including the \opd{} variants, a response without an
answer line is a truncated one and a response that stops has an answer line, up to a few responses per 500 that
give an answer and then continue to the budget; the exceptions are the two \sft{} data controls, which stop
without an answer on 70 (unfiltered) and 10 (complete-only) of 500 responses at 300 problems and on 7 and 0.5 at
900 (Section~\ref{sec:geo}). Table~\ref{tab:teachercond} splits each
student's completion rate by whether the greedy teacher concludes the same test problem, and
Table~\ref{tab:decomp} decomposes each paired accuracy difference by which method concluded a problem.

\begin{table}[H]
  \centering
  \footnotesize
  \caption{\geo, mean counts per 500 test responses over four seeds (single models for the zero-shot rows).
  ``Stops'' means not truncated by the re-tokenisation criterion; ``at budget'' means truncated.}
  \label{tab:crosstab}
  \begin{tabular}{llrrrr}
    \toprule
    & & \multicolumn{2}{c}{Answer line} & \multicolumn{2}{c}{No answer line} \\
    \cmidrule(lr){3-4} \cmidrule(lr){5-6}
    Model & Budget & stops & at budget & at budget & stops \\
    \midrule
    Student, zero-shot & 1{,}536 & 324 & 0 & 175 & 1 \\
    Teacher, zero-shot & 1{,}536 & 391 & 0 & 109 & 0 \\
    Student, zero-shot & 4{,}096 & 325 & 0 & 174 & 1 \\
    Teacher, zero-shot & 4{,}096 & 422 & 0 & 78 & 0 \\
    \sft, 100 & 1{,}536 & 387.5 & 1.2 & 109.2 & 2.0 \\
    \opd, 100 & 1{,}536 & 317.2 & 0.5 & 181.8 & 0.5 \\
    \sft, 300 & 1{,}536 & 363.2 & 0.5 & 134.5 & 1.8 \\
    \opd, 300 & 1{,}536 & 308.0 & 0.5 & 191.0 & 0.5 \\
    \sft, 900 & 1{,}536 & 378.2 & 1.8 & 119.2 & 0.8 \\
    \opd, 900 & 1{,}536 & 313.2 & 0.8 & 186.0 & 0.0 \\
    \sft, 2101 & 1{,}536 & 378.0 & 1.0 & 120.8 & 0.2 \\
    \opd, 2101 & 1{,}536 & 321.0 & 0.2 & 178.5 & 0.2 \\
    \sft{} unfiltered, 300 & 1{,}536 & 310.2 & 0.0 & 120.2 & 69.5 \\
    \sft{} unfiltered, 900 & 1{,}536 & 346.0 & 0.2 & 146.5 & 7.2 \\
    \opd-term, 900 & 1{,}536 & 318.8 & 0.5 & 180.2 & 0.5 \\
    \opd-t07, 900 & 1{,}536 & 307.0 & 1.0 & 191.2 & 0.8 \\
    \opd-fwd, 900 & 1{,}536 & 299.8 & 2.5 & 197.2 & 0.5 \\
        \opd-sftinit, 900 & 1{,}536 & 311.0 & 0.2 & 188.5 & 0.2 \\
    \opd-seqmean, 900 & 1{,}536 & 310.0 & 0.5 & 189.5 & 0.0 \\
    \sft{} complete-only, 300 & 1{,}536 & 357.8 & 0.8 & 131.2 & 10.2 \\
    \sft{} complete-only, 900 & 1{,}536 & 368.8 & 0.8 & 130.0 & 0.5 \\
    \sft, 900 & 4{,}096 & 449.0 & 1.2 & 48.5 & 1.2 \\
    \opd, 900 & 4{,}096 & 331.5 & 0.0 & 167.8 & 0.8 \\
    \bottomrule
  \end{tabular}
\end{table}

\begin{table}[H]
  \centering
  \small
  \caption{\geo, 900 problems: each student's completion rate on the 391 test problems the greedy teacher
  concludes within 1{,}536 tokens and on the 109 it does not (mean over four seeds; the zero-shot student is one
  model). At 4{,}096 tokens the teacher concludes 422 problems.}
  \label{tab:teachercond}
  \begin{tabular}{lccc}
    \toprule
    Student & Overall & Teacher concludes & Teacher does not \\
    \midrule
    Zero-shot & $0.648$ & $0.731$ & $0.349$ \\
    \sft & $0.760$ & $0.848$ & $0.445$ \\
    \opd & $0.628$ & $0.748$ & $0.197$ \\
    \opd-term & $0.638$ & $0.748$ & $0.245$ \\
    \opd-fwd & $0.605$ & $0.721$ & $0.186$ \\
    \opd-sftinit & $0.623$ & $0.749$ & $0.167$ \\
    \sft{} at 4{,}096 tokens & $0.900$ & $0.940$ & $0.686$ \\
    \opd{} at 4{,}096 tokens & $0.663$ & $0.730$ & $0.298$ \\
    \bottomrule
  \end{tabular}
\end{table}

\begin{table}[H]
  \centering
  \footnotesize
  \setlength{\tabcolsep}{4pt}
  \caption{\geo: decomposition of the paired accuracy difference (percentage points of the 500 test problems,
  mean over four seeds) by which method concluded a problem. A problem only one method concludes counts against
  the other, which scores it wrong; $n$ is the mean number of problems in each group. The three components
  and the total are rounded independently, so the sum of the components may differ from the total by $0.1$.}
  \label{tab:decomp}
  \begin{tabular}{lcccc}
    \toprule
    Comparison & Both conclude ($n$) & Only first ($n$) & Only second ($n$) & Total \\
    \midrule
    \opd{} $-$ \sft, 100   & $+0.6$ (287) & $+3.5$ (31) & $-6.0$ (102) & $-1.9$ \\
    \opd{} $-$ \sft, 300   & $+2.4$ (278) & $+3.4$ (30) & $-5.9$ (86)  & $-0.1$ \\
    \opd{} $-$ \sft, 900   & $-1.4$ (285) & $+2.2$ (29) & $-8.8$ (95)  & $-7.9$ \\
    \opd{} $-$ \sft, 2101  & $-2.3$ (291) & $+2.5$ (30) & $-6.8$ (88)  & $-6.7$ \\
    \opd{} $-$ \sft, 900 (4{,}096 tok.) & $-2.5$ (318) & $+1.1$ (13) & $-9.4$ (132) & $-10.8$ \\
    \opd-sftinit $-$ \sft, 900 & $-0.3$ (287) & $+2.2$ (24) & $-7.8$ (93) & $-5.9$ \\
    unfilt.\ $-$ filt.\ \sft, 300 & $+1.8$ (275) & $+3.5$ (35) & $-7.1$ (89) & $-1.8$ \\
    unfilt.\ $-$ filt.\ \sft, 900 & $-0.9$ (307) & $+3.2$ (40) & $-6.1$ (73) & $-3.6$ \\
    \bottomrule
  \end{tabular}
\end{table}

\section{Generation budget}
\label{app:budget}

Table~\ref{tab:budget} re-scores the same four 900-problem models per method at 4{,}096 generated tokens. The
1{,}536 column is a separate evaluation pass of the same weights as Table~\ref{tab:geo}; the four-seed means
differ by $0.1$ (\sft) and $0.4$ (\opd) points and individual seeds by up to $2.8$ points, because greedy
decoding of long solutions is not bit-reproducible across hardware and batch composition.

\begin{table}[H]
  \centering
  \footnotesize
  \caption{\geo, 900 problems, the same four models per method scored at two generation budgets (means with
  question-clustered intervals). The \sft{} minus \opd{} gap grows with the budget because \opd{} students do
  not finish when given more room.}
  \label{tab:budget}
  \begin{tabular}{lcccc}
    \toprule
    & \multicolumn{2}{c}{1{,}536 tokens} & \multicolumn{2}{c}{4{,}096 tokens} \\
    \cmidrule(lr){2-3} \cmidrule(lr){4-5}
    Model & Accuracy & Completion & Accuracy & Completion \\
    \midrule
    Student, zero-shot & $0.366$ & $0.65$ & $0.366$ & $0.65$ \\
    Teacher, zero-shot & $0.570$ & $0.78$ & $0.590$ & $0.84$ \\
    \sft, four seeds & $0.482$ [0.446, 0.520] & $0.76$ & $0.522$ [0.486, 0.559] & $0.90$ \\
    \opd, four seeds & $0.406$ [0.369, 0.445] & $0.63$ & $0.414$ [0.378, 0.452] & $0.66$ \\
    \opd{} $-$ \sft{} (paired) & \multicolumn{2}{c}{$-0.076$ [$-0.102$, $-0.051$]} & \multicolumn{2}{c}{$-0.108$ [$-0.135$, $-0.082$]} \\
    $t$ over seeds & \multicolumn{2}{c}{[$-0.113$, $-0.039$]} & \multicolumn{2}{c}{[$-0.128$, $-0.089$]} \\
    \bottomrule
  \end{tabular}
\end{table}

\section{Control and variants at 900 and 300 problems}
\label{app:control300}

\begin{table}[H]
  \centering
  \scriptsize
  \setlength{\tabcolsep}{3pt}
  \caption{\geo, 900 problems, four seeds each (300-problem rows in Appendix~\ref{app:control300}): the three
  \sft{} data sets (filtered: correct teacher samples, the reference; unfiltered: every teacher sample;
  complete-only: the samples that contain an answer line, correct or not) and plain \opd{} with its five
  variants (defined in Section~\ref{sec:geo} and Appendix~\ref{app:hparams}). Paired differences against the
  filtered \sft, with clustered and seed-level $t$ intervals; bold: intervals that exclude zero.}
  \label{tab:control}
  \begin{tabular}{llcccc}
    \toprule
    & & & & \multicolumn{2}{c}{vs.\ filtered \sft{}} \\
    \cmidrule(lr){5-6}
    & Variant & Accuracy & Completion & paired, clustered 95\% & $t$ over seeds \\
    \midrule
    \multirow{3}{*}{\sft} & filtered & $0.481$ (.022) & $0.76$ & & \\
    & unfiltered & $0.445$ (.013) & $0.69$ & \tb{$-0.036$ [$-0.062$, $-0.012$]} & [$-0.091$, $+0.018$] \\
    & complete-only & $0.473$ (.012) & $0.74$ & $-0.008$ [$-0.032$, $+0.016$] & [$-0.037$, $+0.021$] \\
    \midrule
    \multirow{6}{*}{\opd} & plain & $0.402$ (.006) & $0.63$ & \tb{$-0.079$ [$-0.105$, $-0.053$]} & \tb{[$-0.110$, $-0.048$]} \\
    & term & $0.404$ (.007) & $0.64$ & \tb{$-0.077$ [$-0.101$, $-0.052$]} & \tb{[$-0.112$, $-0.042$]} \\
    & t07 & $0.408$ (.012) & $0.62$ & \tb{$-0.073$ [$-0.099$, $-0.047$]} & \tb{[$-0.109$, $-0.037$]} \\
    & fwd & $0.394$ (.009) & $0.60$ & \tb{$-0.088$ [$-0.113$, $-0.063$]} & \tb{[$-0.132$, $-0.044$]} \\
    & sftinit & $0.422$ (.015) & $0.62$ & \tb{$-0.059$ [$-0.082$, $-0.038$]} & \tb{[$-0.109$, $-0.010$]} \\
    & seqmean & $0.407$ (.014) & $0.62$ & \tb{$-0.075$ [$-0.101$, $-0.050$]} & \tb{[$-0.128$, $-0.022$]} \\
    \bottomrule
  \end{tabular}
\end{table}

\begin{table}[H]
  \centering
  \footnotesize
  \setlength{\tabcolsep}{5pt}
  \caption{\geo, 300 problems, four seeds each: the three \sft{} data sets, plain \opd, and \opd-term, as in
  Table~\ref{tab:control}. No interval excludes zero.}
  \label{tab:control300}
  \begin{tabular}{llcccc}
    \toprule
    & & & & \multicolumn{2}{c}{vs.\ filtered \sft{}} \\
    \cmidrule(lr){5-6}
    & Variant & Accuracy & Completion & paired, clustered 95\% & $t$ over seeds \\
    \midrule
    \multirow{3}{*}{\sft} & filtered & $0.420$ (.013) & $0.73$ & & \\
    & unfiltered & $0.402$ (.016) & $0.62$ & $-0.018$ [$-0.042$, $+0.007$] & [$-0.064$, $+0.028$] \\
    & complete-only & $0.432$ (.011) & $0.72$ & $+0.012$ [$-0.010$, $+0.036$] & [$-0.021$, $+0.046$] \\
    \midrule
    \multirow{2}{*}{\opd} & plain & $0.419$ (.019) & $0.62$ & $-0.001$ [$-0.024$, $+0.022$] & [$-0.020$, $+0.019$] \\
    & term & $0.401$ (.009) & $0.62$ & $-0.019$ [$-0.040$, $+0.003$] & [$-0.039$, $+0.001$] \\
    \bottomrule
  \end{tabular}
\end{table}

\section{Out-of-distribution results and the 32B teacher}
\label{app:ood}

With Qwen3-VL-32B-Instruct as the teacher on 300 \chartqa{} questions (four seeds, same protocol), the teacher's
zero-shot accuracy is $0.858$ (completion $0.986$) against $0.844$ for the 8B, a paired difference of $+1.4$
[$-0.8$, $+3.6$]; its student reaches $0.812$ (sd $0.009$; per seed $0.814/0.812/0.822/0.800$). The per-token
KL to the 32B teacher starts at $0.59$ against $0.40$ to $0.45$ for the 8B and is still $0.23$ against $0.17$ to
$0.18$ at steps 91--100, while the answer-line concentration collapses as in Figure~\ref{fig:dynamics}; we read
this as a description rather than an explanation, since a KL between different teachers is not a distance on a
common scale. The 4B student was not run: it is already at the 8B teacher's level zero-shot ($0.836$ against
$0.844$).

\begin{table}[H]
  \centering
  \footnotesize
  \setlength{\tabcolsep}{3pt}
  \caption{Out-of-distribution accuracy of \chartqa-trained models (four seeds at every budget) on CharXiv
  (461 numeric questions) and ChartQAPro (500 factoid questions), with paired \opd{} minus \sft{} and its
  question-clustered interval. Zero-shot student $0.258$ / $0.276$, teacher $0.412$ / $0.406$.}
  \label{tab:ood}
  \begin{tabular}{lcccc}
    \toprule
    & \multicolumn{2}{c}{CharXiv} & \multicolumn{2}{c}{ChartQAPro} \\
    \cmidrule(lr){2-3} \cmidrule(lr){4-5}
    Questions & \sft{} / \opd & \opd{} $-$ \sft & \sft{} / \opd & \opd{} $-$ \sft \\
    \midrule
    100 & $0.258$ / $0.288$ & $+0.030$ [$+0.005$, $+0.054$] & $0.301$ / $0.306$ & $+0.005$ [$-0.016$, $+0.025$] \\
    300 & $0.278$ / $0.285$ & $+0.007$ [$-0.016$, $+0.030$] & $0.324$ / $0.317$ & $-0.007$ [$-0.027$, $+0.015$] \\
    900 & $0.286$ / $0.290$ & $+0.003$ [$-0.020$, $+0.027$] & $0.317$ / $0.322$ & $+0.005$ [$-0.017$, $+0.026$] \\
    3000 & $0.289$ / $0.287$ & $-0.001$ [$-0.024$, $+0.022$] & $0.326$ / $0.321$ & $-0.005$ [$-0.025$, $+0.014$] \\
    \bottomrule
  \end{tabular}
\end{table}

\section{Sensitivity of the intervals to the resampling unit}
\label{app:intervals}

Tables~\ref{tab:intervals} and \ref{tab:intervalsgeo} give, for every headline comparison, the four per-seed paired differences and three
95\% intervals: \emph{questions}, the question-clustered bootstrap used throughout the paper, which resamples
questions and keeps every seed's outcome on a question together, treating the four trained models as fixed;
\emph{two-way}, a bootstrap that resamples seeds with replacement as well as questions; and \emph{seeds}, the
$t$ interval over the four per-seed differences. A bootstrap that resamples the $4 \times 500$ per-question
outcomes as if they were independent gives intervals about a quarter narrower than the question-clustered ones
(for example [$+1.3$, $+4.2$] against [$+0.8$, $+4.7$] for \opd{} minus \sft{} at 300 \chartqa{} questions) and
is not used. The \geo{} losses at 900 and 2{,}101 problems, at both generation budgets, exclude zero under every
convention; most of the \chartqa{} differences do not.

\begin{table}[H]
  \centering
  \footnotesize
  \setlength{\tabcolsep}{2pt}
    \caption{\chartqa{} and out-of-distribution comparisons: paired differences in percentage points (seeds
  1/2/3/42) and their intervals under three conventions.}
  \label{tab:intervals}
  \begin{tabular}{p{2.4cm}crccc}
    \toprule
    Comparison & Per seed & Mean & Questions & Two-way & Seeds \\
    \midrule
    \chartqa{} 100, \opd{} $-$ \sft & ${+}3.2/{+}1.2/{+}3.0/{+}0.0$ & $+1.8$ & [$-0.1$, $+3.8$] & [$-0.8$, $+4.5$] & [$-0.6$, $+4.3$] \\
    \chartqa{} 300, \opd{} $-$ \sft & ${+}0.6/{+}2.0/{+}2.4/{+}5.8$ & $+2.7$ & [$+0.8$, $+4.7$] & [$-0.1$, $+5.7$] & [$-0.8$, $+6.2$] \\
    \chartqa{} 900, \opd{} $-$ \sft & ${+}1.4/{+}0.8/{+}2.6/{+}1.2$ & $+1.5$ & [$-0.4$, $+3.5$] & [$-0.7$, $+3.9$] & [$+0.3$, $+2.7$] \\
    \chartqa{} 3000, \opd{} $-$ \sft & ${-}1.2/{-}3.2/{-}0.6/{-}1.0$ & $-1.5$ & [$-3.2$, $+0.2$] & [$-3.8$, $+0.7$] & [$-3.3$, $+0.3$] \\
    \chartqa{} 100, \opd{} $-$ \sft{} 38 steps & ${+}2.6/{+}1.6/{+}1.0/{+}0.6$ & $+1.5$ & [$-0.5$, $+3.5$] & [$-0.9$, $+3.8$] & [$+0.1$, $+2.8$] \\
    \chartqa{} 300, \opd{} $-$ \sft{} 38 steps & ${+}1.2/{+}0.6/{+}0.6/{+}1.8$ & $+1.1$ & [$-0.6$, $+2.8$] & [$-0.9$, $+3.0$] & [$+0.1$, $+2.0$] \\
    \chartqa, \opd{} 300 $-$ \sft{} 3000 & ${-}2.6/{-}2.0/{-}1.0/{+}0.2$ & $-1.4$ & [$-3.1$, $+0.4$] & [$-3.6$, $+0.9$] & [$-3.3$, $+0.6$] \\
    \chartqa{} 300, 32B $-$ 8B teacher & ${-}0.6/{-}1.4/{-}0.2/{-}3.6$ & $-1.5$ & [$-3.1$, $+0.2$] & [$-3.8$, $+0.9$] & [$-3.9$, $+1.0$] \\
        CharXiv 100, \opd{} $-$ \sft & ${+}1.5/{+}3.7/{+}3.5/{+}3.3$ & $+3.0$ & [$+0.5$, $+5.4$] & [$-0.1$, $+6.0$] & [$+1.4$, $+4.6$] \\
    \bottomrule
  \end{tabular}
\end{table}

\begin{table}[H]
  \centering
  \footnotesize
  \setlength{\tabcolsep}{2pt}
  \caption{\geo{} comparisons, as in Table~\ref{tab:intervals}.}
  \label{tab:intervalsgeo}
  \begin{tabular}{p{2.4cm}crccc}
    \toprule
    Comparison & Per seed & Mean & Questions & Two-way & Seeds \\
    \midrule
    \geo{} 100, \opd{} $-$ \sft & ${+}0.0/{+}0.2/{-}2.6/{-}5.2$ & $-1.9$ & [$-4.5$, $+0.7$] & [$-5.6$, $+1.8$] & [$-5.9$, $+2.1$] \\
    \geo{} 300, \opd{} $-$ \sft & ${+}0.8/{-}1.8/{+}0.8/{+}0.0$ & $-0.1$ & [$-2.4$, $+2.2$] & [$-3.0$, $+2.9$] & [$-2.0$, $+1.9$] \\
    \geo{} 900, \opd{} $-$ \sft & ${-}6.6/{-}6.4/{-}10.6/{-}8.0$ & $-7.9$ & [$-10.5$, $-5.3$] & [$-11.4$, $-4.4$] & [$-11.0$, $-4.8$] \\
    \geo{} 2101, \opd{} $-$ \sft & ${-}5.2/{-}9.0/{-}5.6/{-}7.0$ & $-6.7$ & [$-9.2$, $-4.3$] & [$-10.0$, $-3.5$] & [$-9.4$, $-4.0$] \\
    \geo{} 900 (4{,}096 tok.), \opd{} $-$ \sft & ${-}11.2/{-}10.8/{-}9.2/{-}12.2$ & $-10.8$ & [$-13.5$, $-8.2$] & [$-14.1$, $-7.5$] & [$-12.8$, $-8.9$] \\
    \geo{} 300, unfiltered $-$ filtered \sft & ${-}5.2/{-}1.8/{+}1.8/{-}2.0$ & $-1.8$ & [$-4.2$, $+0.7$] & [$-5.6$, $+2.1$] & [$-6.4$, $+2.8$] \\
    \geo{} 900, unfiltered $-$ filtered \sft & ${-}1.4/{-}1.2/{-}8.6/{-}3.4$ & $-3.6$ & [$-6.2$, $-1.2$] & [$-8.0$, $+0.3$] & [$-9.1$, $+1.8$] \\
    \geo{} 900, \opd-sftinit $-$ \sft & ${-}3.2/{-}5.4/{-}10.4/{-}4.8$ & $-5.9$ & [$-8.2$, $-3.8$] & [$-10.0$, $-2.4$] & [$-10.9$, $-1.0$] \\
    \geo{} 900, \opd-sftinit $-$ \opd & ${+}3.4/{+}1.0/{+}0.2/{+}3.2$ & $+1.9$ & [$-0.2$, $+4.2$] & [$-1.1$, $+5.0$] & [$-0.6$, $+4.5$] \\
        \geo{} 900, \opd-fwd $-$ \opd & ${-}1.4/{+}0.6/{-}1.8/{-}1.0$ & $-0.9$ & [$-2.9$, $+1.1$] & [$-3.6$, $+1.8$] & [$-2.6$, $+0.8$] \\
    \geo{} 300, complete-only $-$ filtered \sft & ${-}0.4/{+}1.8/{+}4.0/{-}0.4$ & $+1.2$ & [$-1.0$, $+3.6$] & [$-2.0$, $+4.7$] & [$-2.1$, $+4.6$] \\
    \geo{} 900, complete-only $-$ filtered \sft & ${-}1.4/{+}1.6/{-}2.8/{-}0.6$ & $-0.8$ & [$-3.2$, $+1.6$] & [$-4.1$, $+2.5$] & [$-3.7$, $+2.1$] \\
    \geo{} 900, complete-only $-$ unfiltered \sft & ${+}0.0/{+}2.8/{+}5.8/{+}2.8$ & $+2.9$ & [$+0.7$, $+5.0$] & [$-0.6$, $+6.1$] & [$-0.9$, $+6.6$] \\
    \geo{} 900, \opd-seqmean $-$ \opd & ${+}0.0/{+}2.6/{-}1.2/{+}0.2$ & $+0.4$ & [$-1.4$, $+2.1$] & [$-2.2$, $+3.2$] & [$-2.1$, $+2.9$] \\
    \chartqa{} 300, \opd-seqmean $-$ \opd & ${-}0.2/{+}0.0/{+}1.0/{-}1.4$ & $-0.1$ & [$-1.4$, $+1.1$] & [$-1.9$, $+1.6$] & [$-1.7$, $+1.4$] \\
    \bottomrule
  \end{tabular}
\end{table}

\section{Numeric tolerance}
\label{app:tolerance}

Relaxed accuracy accepts a numeric answer within 5\% of the gold value. Table~\ref{tab:tolerance} re-scores
every \geo{} model from its stored predictions at 1\% and 0.1\%. Absolute accuracies fall by 1 to 4 points; the
ordering of the methods is unchanged, and every interval that excludes zero at 5\% does so at 1\% and 0.1\% as
well, with endpoints moving by up to a point.

\begin{table}[H]
  \centering
  \small
  \caption{\geo{} accuracy (mean over four seeds) at three numeric tolerances, with the paired \opd{} minus
  \sft{} difference in points and its question-clustered interval. Zero-shot student / teacher: $0.366$ / $0.570$
  at 5\%, $0.354$ / $0.552$ at 1\%, $0.330$ / $0.534$ at 0.1\%.}
  \label{tab:tolerance}
  \begin{tabular}{llccc}
    \toprule
    Tolerance & Problems & \sft & \opd & \opd{} $-$ \sft \\
    \midrule
    \multirow{4}{*}{5\%}   & 100  & $0.423$ & $0.404$ & $-1.9$ [$-4.5$, $+0.7$] \\
                           & 300  & $0.420$ & $0.419$ & $-0.1$ [$-2.4$, $+2.2$] \\
                           & 900  & $0.481$ & $0.402$ & $-7.9$ [$-10.5$, $-5.3$] \\
                           & 2101 & $0.478$ & $0.411$ & $-6.7$ [$-9.2$, $-4.3$] \\
    \midrule
    \multirow{4}{*}{1\%}   & 100  & $0.407$ & $0.395$ & $-1.2$ [$-3.6$, $+1.3$] \\
                           & 300  & $0.408$ & $0.406$ & $-0.2$ [$-2.5$, $+2.1$] \\
                           & 900  & $0.476$ & $0.392$ & $-8.5$ [$-11.2$, $-5.9$] \\
                           & 2101 & $0.466$ & $0.397$ & $-6.8$ [$-9.3$, $-4.5$] \\
    \midrule
    \multirow{4}{*}{0.1\%} & 100  & $0.396$ & $0.377$ & $-1.9$ [$-4.5$, $+0.7$] \\
                           & 300  & $0.395$ & $0.386$ & $-0.8$ [$-3.2$, $+1.5$] \\
                           & 900  & $0.459$ & $0.374$ & $-8.4$ [$-10.9$, $-5.9$] \\
                           & 2101 & $0.449$ & $0.379$ & $-7.1$ [$-9.5$, $-4.7$] \\
    \bottomrule
  \end{tabular}
\end{table}

\section{\chartqa{} variants}
\label{app:chartqa-variants}

Table~\ref{tab:variants} scores the \opd{} variants of Section~\ref{sec:geo} on \chartqa, where the teacher
concludes 99\% of its solutions.

\begin{table}[H]
  \centering
  \footnotesize
  \caption{\opd{} variants on \chartqa{} (four seeds each, across-seed sd in brackets): no variant differs from
  plain \opd. Reference points: \sft{} $0.793$ / $0.799$ and \opd{} $0.811$ / $0.826$ at 100 / 300 questions.
  Intervals are question-clustered.}
  \label{tab:variants}
  \setlength{\tabcolsep}{4pt}
  \begin{tabular}{llccc}
    \toprule
    Questions & Variant & Accuracy & vs.\ \sft{} (paired) & vs.\ \opd{} (paired) \\
    \midrule
    100 & \opd-term & $0.814$ (.006) & $+0.021$ [$+0.001$, $+0.042$] & $+0.002$ [$-0.011$, $+0.015$] \\
    300 & \opd-term & $0.823$ (.011) & $+0.024$ [$+0.004$, $+0.042$] & $-0.004$ [$-0.016$, $+0.009$] \\
    300 & \opd-fwd & $0.828$ (.005) & $+0.029$ [$+0.009$, $+0.048$] & $+0.001$ [$-0.011$, $+0.015$] \\
    300 & \opd-sftinit & $0.822$ (.005) & $+0.023$ [$+0.008$, $+0.038$] & $-0.004$ [$-0.021$, $+0.012$] \\
    300 & \opd-seqmean & $0.825$ (.007) & $+0.025$ [$+0.007$, $+0.044$] & $-0.001$ [$-0.014$, $+0.011$] \\
    \bottomrule
  \end{tabular}
\end{table}

\section{Training length of \sft{} on \chartqa}
\label{app:sft-length}

Table~\ref{tab:sftlength} gives the \sft{} training-length control of Section~\ref{sec:chartqa}.

\begin{table}[H]
  \centering
  \footnotesize
  \setlength{\tabcolsep}{4pt}
  \caption{\sft{} trained for fewer steps than the protocol's 302, four seeds each (across-seed sd in
  brackets), and the paired \opd{} minus \sft{} difference against each length (question-clustered intervals).
  150 steps of 16 sequences equals \opd's number of training sequences. The best length was selected on the
  test set.}
  \label{tab:sftlength}
  \begin{tabular}{lcccc}
    \toprule
    & \multicolumn{2}{c}{100 questions} & \multicolumn{2}{c}{300 questions} \\
    \cmidrule(lr){2-3} \cmidrule(lr){4-5}
    \sft{} steps & Accuracy & \opd{} $-$ \sft & Accuracy & \opd{} $-$ \sft \\
    \midrule
    19 & $0.791$ (.007) & $+0.021$ [$+0.001$, $+0.039$] & $0.802$ (.009) & $+0.024$ [$+0.008$, $+0.040$] \\
    38 & $0.797$ (.012) & $+0.015$ [$-0.005$, $+0.035$] & $0.816$ (.005) & $+0.011$ [$-0.006$, $+0.028$] \\
    75 & $0.788$ (.005) & $+0.024$ [$+0.003$, $+0.044$] & $0.811$ (.005) & $+0.016$ [$-0.002$, $+0.034$] \\
    150 & $0.788$ (.009) & $+0.024$ [$+0.003$, $+0.043$] & $0.808$ (.004) & $+0.018$ [$-0.001$, $+0.039$] \\
    302 (protocol) & $0.793$ (.022) & $+0.018$ [$-0.001$, $+0.038$] & $0.799$ (.015) & $+0.027$ [$+0.008$, $+0.047$] \\
    \bottomrule
  \end{tabular}
\end{table}

\section{Data-scaling slopes}
\label{app:scaling}

Table~\ref{tab:scaling} fits accuracy against $\log_{10}$ of the training budget, per seed, over the four
budgets of Figure~\ref{fig:curves}. \sft{} gains 3 to 5 points per decade of questions; \opd{} gains none, because
its compute is fixed at 150 steps of 16 rollouts, or 2{,}400 rollouts, which is 24 passes over 100 questions, 8
over 300, 2.7 over 900, and 0.8 over 3{,}000. The slopes describe this protocol: with 1.5 decades of data and one
model size they are not scaling laws, and whether \opd's slope rises when its step count grows with the data is
the experiment the protocol leaves open, which we leave to future work.

\begin{table}[H]
  \centering
  \footnotesize
  \caption{Accuracy (mean over four seeds) at the four budgets and the slope of a per-seed least-squares fit of
  accuracy on $\log_{10}$(budget), in points per decade, with the $t$ interval over the four seeds and the $R^2$
  of the fit to the seed means.}
  \label{tab:scaling}
  \begin{tabular}{llcccccc}
    \toprule
    Task & Method & 100 & 300 & 900 & 3000 / 2101 & Slope (pts / decade) & $R^2$ \\
    \midrule
    \chartqa & \sft & $0.793$ & $0.799$ & $0.808$ & $0.840$ & $+3.1$ [$+1.3$, $+4.8$] & 0.87 \\
    \chartqa & \opd & $0.811$ & $0.826$ & $0.823$ & $0.825$ & $+0.8$ [$-0.9$, $+2.4$] & 0.51 \\
    \geo & \sft & $0.423$ & $0.420$ & $0.481$ & $0.478$ & $+5.1$ [$+2.7$, $+7.6$] & 0.78 \\
    \geo & \opd & $0.404$ & $0.419$ & $0.402$ & $0.411$ & $+0.0$ [$-2.7$, $+2.8$] & 0.00 \\
    \bottomrule
  \end{tabular}
\end{table}

\section{Per-seed results}
\label{app:seeds}

Table~\ref{tab:seeds} lists the accuracy of every trained model behind the paper's means, seed by seed; the
unfiltered and complete-only \sft{} controls, the \opd{} variants, and the re-scored evaluations are included.

\begin{table}[H]
  \centering
  \footnotesize
  \setlength{\tabcolsep}{3pt}
  \caption{Per-seed accuracy on the 500 test questions (seeds 1, 2, 3, and 42) and the four-seed mean.}
  \label{tab:seeds}
  \begin{tabular}{lllccccc}
    \toprule
    & & & \multicolumn{4}{c}{Seed} & \\
    \cmidrule(lr){4-7}
    Task & Budget & Method & 1 & 2 & 3 & 42 & Mean \\
    \midrule
    \chartqa & 100 questions & \sft & $0.778$ & $0.808$ & $0.770$ & $0.816$ & $0.793$ \\
     &  & \opd & $0.810$ & $0.820$ & $0.800$ & $0.816$ & $0.811$ \\
     & 300 questions & \sft & $0.814$ & $0.806$ & $0.800$ & $0.778$ & $0.799$ \\
     &  & \opd & $0.820$ & $0.826$ & $0.824$ & $0.836$ & $0.826$ \\
     &  & \opd, 32B teacher & $0.814$ & $0.812$ & $0.822$ & $0.800$ & $0.812$ \\
     &  & \opd-fwd & $0.832$ & $0.822$ & $0.826$ & $0.832$ & $0.828$ \\
     &  & \opd-sftinit & $0.830$ & $0.820$ & $0.820$ & $0.820$ & $0.822$ \\
     &  & \opd-seqmean & $0.818$ & $0.826$ & $0.834$ & $0.822$ & $0.825$ \\
     & 900 questions & \sft & $0.814$ & $0.818$ & $0.794$ & $0.808$ & $0.808$ \\
     &  & \opd & $0.828$ & $0.826$ & $0.820$ & $0.820$ & $0.823$ \\
     & 3{,}000 questions & \sft & $0.846$ & $0.846$ & $0.834$ & $0.834$ & $0.840$ \\
     &  & \opd & $0.834$ & $0.814$ & $0.828$ & $0.824$ & $0.825$ \\
    \midrule
    \geo & 100 problems & \sft & $0.394$ & $0.424$ & $0.420$ & $0.456$ & $0.423$ \\
     &  & \opd & $0.394$ & $0.426$ & $0.394$ & $0.404$ & $0.404$ \\
     & 300 problems & \sft & $0.436$ & $0.418$ & $0.404$ & $0.422$ & $0.420$ \\
     &  & \sft{} (complete-only) & $0.432$ & $0.436$ & $0.444$ & $0.418$ & $0.432$ \\
     &  & \opd & $0.444$ & $0.400$ & $0.412$ & $0.422$ & $0.419$ \\
     & 900 problems & \sft & $0.472$ & $0.464$ & $0.514$ & $0.476$ & $0.481$ \\
     &  & \sft{} (unfiltered) & $0.458$ & $0.452$ & $0.428$ & $0.442$ & $0.445$ \\
     &  & \sft{} (complete-only) & $0.458$ & $0.480$ & $0.486$ & $0.470$ & $0.473$ \\
     &  & \opd & $0.406$ & $0.400$ & $0.408$ & $0.396$ & $0.402$ \\
     &  & \opd-term & $0.400$ & $0.412$ & $0.408$ & $0.398$ & $0.404$ \\
     &  & \opd-t07 & $0.392$ & $0.408$ & $0.412$ & $0.422$ & $0.408$ \\
     &  & \opd-fwd & $0.392$ & $0.406$ & $0.390$ & $0.386$ & $0.394$ \\
     &  & \opd-sftinit & $0.440$ & $0.410$ & $0.410$ & $0.428$ & $0.422$ \\
     &  & \opd-seqmean & $0.406$ & $0.426$ & $0.396$ & $0.398$ & $0.407$ \\
     & 900, 1{,}536-token re-score & \sft & $0.458$ & $0.482$ & $0.486$ & $0.504$ & $0.482$ \\
     &  & \opd & $0.414$ & $0.396$ & $0.410$ & $0.406$ & $0.406$ \\
     & 900, 4{,}096-token re-score & \sft & $0.526$ & $0.518$ & $0.512$ & $0.534$ & $0.522$ \\
     &  & \opd & $0.414$ & $0.410$ & $0.420$ & $0.412$ & $0.414$ \\
     & 2{,}101 problems & \sft & $0.468$ & $0.482$ & $0.478$ & $0.484$ & $0.478$ \\
     &  & \opd & $0.416$ & $0.392$ & $0.422$ & $0.414$ & $0.411$ \\
    \bottomrule
  \end{tabular}
\end{table}

\section{Engineering notes}
\label{app:engineering}

Training an 8B teacher and a 2B student together on one 40 GB GPU with full-vocabulary KL required: left-padded
prompts so that generation positions align across the batch; \texttt{logits\_to\_keep} so that neither model
materialises logits over the prompt or image tokens; the KL computed in fp32 in chunks over the sequence;
micro-batching of the scoring pass with gradient accumulation; and rollouts at batch 16 (the rollout is 90\% of
step time and doubling its batch doubled throughput at constant peak memory of 27 GB). Checkpoints (adapter,
optimizer, scheduler, data cursor) go to a remote repository every 25 steps and a killed job resumes from the
latest one. vLLM \citep{kwon2023vllm} evaluates 500 questions in about 20 seconds. For a 32B teacher the
student is pinned to one GPU and the teacher sharded over two more.

\end{document}